%% file: paper.tex
\documentclass[runningheads,envcountsame]{llncs}
\pdfoutput=1

\usepackage[utf8]{inputenc}
\usepackage{color}
\usepackage[ruled,vlined]{algorithm2e}
\usepackage{tikz}
\usepackage[font=small,labelfont=bf]{caption}
\usepackage{xspace}
\usepackage{paralist}
\usepackage{amsfonts}
\usepackage{amssymb}
\usepackage{amsmath}
\usepackage{stmaryrd}
\usepackage{wasysym}
\usepackage{mathrsfs}
\usepackage{enumitem}
\usepackage{thmtools}
\usepackage{graphicx}
\usepackage{pbox}
\usepackage{multirow}
\usepackage{diagbox}
\usepackage[english]{babel}
\usepackage[colorinlistoftodos]{todonotes}
\usepackage{algpseudocode}
\usepackage{url}
\usepackage{mwe,tikz}\usepackage[percent]{overpic}
\usepackage{comment}
\usepackage{multicol}
\usepackage{xspace}
\usepackage{stmaryrd} 
\usepackage{subfig}
\usepackage{wasysym}
\usepackage{xparse}
\usepackage{graphicx}
\usepackage{colortbl}
\usepackage{ifthen}
\usepackage{adjustbox}          
\usetikzlibrary{calc}
\usepackage{collcell}

\newcommand{\ec}[1]{\makebox[#1][c]{\rule[-2pt]{0pt}{1em}}}
\newlength{\gapw}
\newcounter{val} 
\newcommand*{\MinNumber}{0}%
\newcommand*{\MaxNumber}{14}%
 
\newcommand{\ApplyGradientSmall}[1]{%
        \pgfmathsetmacro{\PercentColor}{100*(#1-\MinNumber)/(\MaxNumber-\MinNumber)}%
	\setcounter{val}{#1}%
	\hspace{-0em}%
	\ifthenelse{\value{val}=20}{\colorbox{white!100!white}{\ec{0.39em}}\hspace*{-0.39em}}{%
	\ifthenelse{\value{val}=14}{\colorbox{black!100!black}{\ec{0.39em}}}{%
        \colorbox{red!\PercentColor!yellow}{\ec{0.39em}}%
	}}%
}
\newcommand{\ApplyGradientNormal}[1]{%
        \pgfmathsetmacro{\PercentColor}{100*(#1-\MinNumber)/(\MaxNumber-\MinNumber)}%
	\setcounter{val}{#1}%
	\hspace{-0em}%
	\ifthenelse{\value{val}=20}{\colorbox{white!100!white}{\ec{0.67em}}\hspace*{-0.5em}}{%
	\ifthenelse{\value{val}=14}{\colorbox{black!100!black}{\ec{0.67em}}}{%
        \colorbox{red!\PercentColor!yellow}{\ec{0.67em}}%
	}}%
}
\newcommand{\ApplyGradientLarge}[1]{%
        \pgfmathsetmacro{\PercentColor}{100*(#1-\MinNumber)/(\MaxNumber-\MinNumber)}%
	\setcounter{val}{#1}%
	\hspace{-0em}%
	\ifthenelse{\value{val}=20}{\colorbox{white!100!white}{\ec{1em}}\hspace*{-0.5em}}{%
	\ifthenelse{\value{val}=14}{\colorbox{black!100!black}{\ec{1em}}}{%
        \colorbox{red!\PercentColor!yellow}{\ec{1em}}%
	}}%
}
 
\newcolumntype{S}{>{\collectcell\ApplyGradientSmall}c<{\endcollectcell}}
\newcolumntype{N}{>{\collectcell\ApplyGradientNormal}c<{\endcollectcell}}
\newcolumntype{L}{>{\collectcell\ApplyGradientLarge}c<{\endcollectcell}}
\newcommand{\lmt}[1]{\makebox[1.9em][r]{$#1$}} 

\let\oldReturn\Return
\definecolor{Black}  {RGB}{0,0,0}

\input{macros}

\input{tik}

\begin{document}
\pagestyle{headings}  

\title{Automated Reasoning in Temporal \DLite}

\author{Sabiha Tahrat\inst{1} \and  German Braun\inst{2} \and 
  Alessandro Artale\inst{3} \and Marco Gario\inst{3} \and \\ Ana Ozaki\inst{4}
}

\authorrunning{Tahrat et. al}

\institute{
Universit\'e de Paris, France \email{sabiha.tahrat@parisdescartes.fr} \and
Universidad Nacional del Comahue, Argentina \email{german.braun@fi.uncoma.edu.ar}
  \and
Free University of Bozen-Bolzano, Italy
  \email{artale@inf.unibz.it, marco.gario@gmail.com} \and
   University of Bergen, Norway
  \email{Ana.Ozaki@uib.no} 
}

\maketitle  

\input{0-abstract.tex}

\input{1-introduction.tex}

\input{2-preliminaries.tex}
\input{3-translation.tex}
\input{4-implementation.tex}

\input{5-evaluation.tex}

\input{6-conclusion.tex}



\newcommand{\SortNoop}[1]{}
\bibliographystyle{abbrv}

\end{document}

%% file: macros.tex
\newcommand{\nb}[1]{}

\SetCommentSty{mycommfont}

\newcommand{\eset}{\emptyset}

\newcommand{\sqs}{\sqsubseteq}

\newcommand{\U}{\ensuremath{\mathbin{\mathcal{U}}}\xspace}
\newcommand{\Since}{\ensuremath{\mathbin{\mathcal{S}}}\xspace}

\newcommand{\Next}{\ensuremath{\ocircle}}
\newcommand{\nxt}{{\ensuremath{\raisebox{0.25ex}{\text{\scriptsize{$\bigcirc$}}}}}}

\newcommand{\D}{\Diamond}

\newcommand{\B}{\Box}

\newcommand{\monodic}{\ensuremath{\mathop{\ooalign{$\Box$ \cr \kern0.55ex \raisebox{0.2ex}{\scalebox{0.5}{$1$}}}\rule{0pt}{1.5ex} \kern-0.7ex}}\xspace}

\newcommand{\OWL}{\textsl{OWL}}
\newcommand{\OWLQL}{\textsl{OWL\,2\,QL}}

\newcommand{\DLite}{\textsl{DL-Lite}\xspace}
\newcommand{\DLites}{\textsl{DL-Lites}\xspace}

\newcommand{\DLitebn}{\ensuremath{\DLite_\textit{bool}^{\cal N}}}

\newcommand{\ALC}{\ensuremath{\smash{\mathcal{ALC}}}\xspace}

\newcommand{\NC}{\ensuremath{{\sf N_C}}\xspace}
\newcommand{\NI}{\ensuremath{{\sf N_I}}\xspace}
\newcommand{\NR}{\ensuremath{{\sf N_R}}\xspace}

\newcommand{\NGl}{\ensuremath{\textsf{N}_{\textsf{G}}\xspace}}
\newcommand{\NL}{\ensuremath{\textsf{N}_{\textsf{L}}\xspace}}

\newcommand{\role}{\ensuremath{\mathsf{role}_{\Kmc}}\xspace}

\newcommand{\inv}{\ensuremath{\mathsf{inv}}\xspace}

\newcommand{\subk}{\ensuremath{\textit{sub}(\K^{\ddagger})}\xspace}
\newcommand{\subphi}{\ensuremath{\textit{sub}(\varphi)}\xspace}

\newcommand{\cl}{\ensuremath{\mathsf{cl}}\xspace}

\renewcommand{\Return}{\State\oldReturn}
\newcommand{\lengthTbox}{\ensuremath{L_{t}}\xspace}
\newcommand{\lengthConcept}{\ensuremath {L_{c}}\xspace}
\newcommand{\MaxQ}{\ensuremath{Q}\xspace}

\newcommand{\LTL}{\ensuremath{\textsl{LTL}}\xspace}

\newcommand{\PTL}{\ensuremath{\textsl{LTL}_P}\xspace}

\newcommand{\QTLO}{\ensuremath{\mathcal{QTL}_{1}}\xspace}

\newcommand{\TDLLite}{\ensuremath{\textsl{TDL-Lite}}\xspace}
\newcommand{\TDLLiteN}{\ensuremath{\textsl{T}^{\Nbb}\textsl{DL-Lite}}\xspace}

\newcommand{\TuDLite}[2]{\ensuremath{\textsl{T}^{#1}_{\U\Since}\DLite_{\textit{#2}}^{\mathcal{N}}}\xspace}

\newcommand{\TdxDLbn}{\ensuremath{\smash{\textsl{T}_{F\!P\!X}\DLitebn}}}

\newcommand{\Rdiamond}{\Diamond_{\!\scriptscriptstyle F}}
\newcommand{\Rbox}{\Box_{\!\scriptscriptstyle F}}
\newcommand{\Rnext}{\nxt_{\!\scriptscriptstyle F}}
\newcommand{\Ldiamond}{\Diamond_{\!\scriptscriptstyle P}}
\newcommand{\Lbox}{\Box_{\!\scriptscriptstyle P}}
\newcommand{\Lnext}{\nxt_{\!\scriptscriptstyle P}}
\newcommand{\SVdiamond}{\mathop{\ooalign{$\Diamond$ \cr \kern0.5ex
    \raisebox{0.35ex}{\scalebox{0.7}{$*$}}} \kern-0.9ex}}
\newcommand{\SVbox}{\mathop{\ooalign{$\Box$ \cr \kern0.42ex
    \raisebox{0.35ex}{\scalebox{0.7}{$*$}}}\rule{0pt}{1.5ex} \kern-0.7ex}}

\newcommand{\PSpace}{\textsc{PSpace}}
\newcommand{\ExpTime}{\textsc{ExpTime}}

\newcommand{\Mmf}{\ensuremath{\mathfrak{M}}\xspace}

\newcommand{\Wmf}{\ensuremath{\mathfrak{W}}\xspace}

\newcommand{\Amc}{\ensuremath{\mathcal{A}}\xspace}
\newcommand{\Bmc}{\ensuremath{\mathcal{B}}\xspace}
\newcommand{\Cmc}{\ensuremath{\mathcal{C}}\xspace}
\newcommand{\Dmc}{\ensuremath{\mathcal{D}}\xspace}

\newcommand{\Kmc}{\ensuremath{\mathcal{K}}\xspace}
\newcommand{\Lmc}{\ensuremath{\mathcal{L}}\xspace}

\newcommand{\Rmc}{\ensuremath{\mathcal{R}}\xspace}
\newcommand{\Smc}{\ensuremath{\mathcal{S}}\xspace}
\newcommand{\Tmc}{\ensuremath{\mathcal{T}}\xspace}
\newcommand{\Umc}{\ensuremath{\mathcal{U}}\xspace}

\newcommand{\Xmc}{\ensuremath{\mathcal{X}}\xspace}

\newcommand{\Nbl}{\ensuremath{\mathbb{N}}\xspace}

\newcommand{\Z}{\mathbb{Z}}
 
\newcommand{\Nbb}{\ensuremath{\mathbb{N}}\xspace}
\newcommand{\Zbb}{\ensuremath{\mathbb{Z}}\xspace}

\newcommand{\A}{\ensuremath{\mathcal{A}}}
\newcommand{\K}{\ensuremath{\mathcal{K}}}
\newcommand{\I}{\ensuremath{\mathcal{I}}}
\newcommand{\T}{\ensuremath{\mathcal{T}}}

\newcommand{\QT}{Q_\T}

\renewcommand{\Return}{\State\oldReturn}
\newcommand{\tdllitefpx}{$T_{FPX}DL$-$Lite^{\mathcal{N}}_{bool}$}

\newcommand{\splus}{\,\raisebox{.25\height}{\scalebox{.8}{+}}\,}
\newcommand{\Aalta}{\texttt{Aalta}\xspace}

\newcommand{\NuXMVBDD}{\texttt{NuXMV-BDD}\xspace}
\newcommand{\NuXMVSBMC}{\texttt{NuXMV-SBMC}\xspace}

\newcommand{\PLTLGraph}{\texttt{pltl} (\texttt{graph})\xspace}
\newcommand{\PLTLTree}{\texttt{pltl} (\texttt{tree})\xspace}
\newcommand{\TRPpp}{\texttt{TRP\protect\raisebox{0.35ex}{\scriptsize{+}{+}}}\xspace}
\newcommand{\TRPppBFS}{\texttt{TRP\protect\raisebox{0.35ex}{\scriptsize{+}{+}}(\texttt{BFS})}\xspace}
\newcommand{\TRPppDFS}{\texttt{TRP\protect\raisebox{0.35ex}{\scriptsize{+}{+}}(\texttt{DFS})}\xspace}
\newcommand{\NuXMVIC}{\texttt{NuXMV-IC3}\xspace}


%% file: tik.tex
\pgfdeclarelayer{edgelayer}
\pgfdeclarelayer{nodelayer}
\pgfsetlayers{edgelayer,nodelayer,main}

\tikzstyle{none}=[inner sep=0pt]
\tikzstyle{invisible}=[inner sep=0pt]
\tikzstyle{rn}=[circle,fill=Red,draw=Black,line width=0.8 pt]
\tikzstyle{gn}=[circle,fill=White,draw=Black,line width=0.8 pt]
\tikzstyle{yn}=[circle,fill=Yellow,draw=Black,line width=0.8 pt]
\tikzstyle{simple}=[circle,fill=White,draw=Black]
\tikzstyle{newstyle1}=[circle,fill=Black,draw=Black,line width=2 pt,inner sep=0pt]
\tikzstyle{simple2}=[-,dashed,draw=Black]
\tikzstyle{simpledotted}=[-,dotted,draw=Black]
\tikzstyle{simple}=[-,draw=Black,line width=2.000]
\tikzstyle{new}=[-,draw=Black,line width=2.000]
\tikzstyle{arrow}=[-,draw=Black,postaction={decorate},decoration={markings,mark=at position .5 with {\arrow{>}}},line width=2.000]
\tikzstyle{tick}=[-,draw=Black,postaction={decorate},decoration={markings,mark=at position .5 with {\draw (0,-0.1) -- (0,0.1);}},line width=2.000]
\tikzstyle{newstyle2}=[-latex,draw=Black]
\tikzstyle{newstyle3}=[->,dotted,draw=Black]
\tikzstyle{newstyle6}=[-latex,draw=Black]
\tikzstyle{newstyle5}=[-latex,draw=red]

%% file: 0-abstract.tex
\begin{abstract} 
  This paper investigates the feasibility of automated reasoning over
  temporal \DLite (\TDLLite) knowledge bases (KBs). We test the usage
  of off-the-shelf \LTL\ reasoners to check satisfiability of \TDLLite
  KBs. In particular, we test the robustness and the scalability of
  reasoners when dealing with \TDLLite TBoxes paired with a temporal
  ABox.
  %
  We conduct various experiments 
  to analyse the performance of different reasoners by randomly
  generating \TDLLite KBs and then measuring the running time
  and the size of the translations.
%
  Furthermore, in an effort to make the usage of \TDLLite KBs a
  reality, we present a fully fledged tool with a graphical interface
  to design them. Our interface is based on conceptual modelling
  principles and it is integrated with 
  our translation tool 
  and a temporal reasoner.
\end{abstract}


%% file: 1-introduction.tex
\section{Introduction}
\label{sec:intro}

Ontology languages are a central research topic in the
Semantic Web community where \OWL\ and its various fragments have been
adopted as a W3C standard ontology language. While \OWL\ allows to
capture and reason over \emph{static} ontologies, here we are
interested in representing and reasoning over \emph{dynamic}
ontologies, i.e., ontologies able to capture the temporal behaviour of
its main primitives (classes, object- and data-properties).

Reasoning over temporal ontologies has been studied in the literature
on temporal representation languages (see~\cite{ArtFra2,LutzWZ08} for
surveys on temporal extensions of description logics).
The complexity of standard reasoning tasks, such as satisfiability
and entailment,
in \ALC-based temporal DLs is known to be hard,
ranging from \ExpTime\ to undecidable~\cite{LutzWZ08,BaaEtAl2}, while it 
becomes easier when extending \DLite-based DLs~\cite{ArtEtAl3}.
There is a growing interest in ontologies based on \DLite,
which is the backbone of the W3C standard ontology language \OWLQL---the
main formalism underpinning the OBDA
paradigm~\cite{PoggiLCGLR08,ArtEtAl1}. While temporal \DLites are 
not yet an \OWL\ standard, they 
have nonetheless been 
studied as a formalism to represent and reason over dynamic
ontologies, in particular when such ontologies have the shape of
conceptual models~\cite{AKRZ:ER10,ArtEtAl3}.


In this paper, we study the practical feasibility of reasoning over
temporal \DLite (\TDLLite) extensions.
In particular, as \TDLLite we consider here the \tdllitefpx\ logic,
the most expressive of the tractable \DLite family combined with \LTL\
allowing for: full Boolean connectives in the construction of
concepts, cardinalities and inverse constraints on roles, \LTL-based
temporal operators applied to
   concepts, distinction between
\emph{global} (i.e., time-invariant) and \emph{local}
roles. Ontologies in \TDLLite\ are expressed via \emph{concept
  inclusion axioms}, so called TBox, intended to express the
constraints of a given application domain via axioms that hold at each
point in time (i.e., a TBox expresses a global knowledge), paired with
temporal assertions (i.e., a so called temporal ABox) expressing
timestamped factual knowledge. The complexity of reasoning over
\TDLLite\ KBs, i.e., a set of TBox and ABox axioms, is known to be
\PSpace-complete~\cite{ArtEtAl3}.



The main purpose here is to collect experimental evidences on the
feasibility of automated reasoning on \TDLLite\ KBs by leveraging on
existing off-the-shelf \LTL solvers.
The key idea is to map such KBs into equisatisfiable \LTL
formulas by applying the mapping described by Artale et
al. (2014)~\cite{ArtEtAl3}. Since \TDLLite\ admits both past and
future operators interpreted over $\Zbb$ while \LTL reasoners deals
just with $\Nbb$, in our study, 
we also consider the simpler logic $\TDLLiteN$ that is able to express
just future temporal formulas interpreted over \Nbb.
Our main contributions are:
\begin{enumerate}[label=(\arabic*)]
\item the development of a non-trivial extension of the tool {\tt
    crowd}~\cite{Braun-KI} to draw \emph{temporal} conceptual schemas,
  featuring an option to populate the schema with timestamped
  instances, which can be automatically mapped into \TDLLite\ KBs;
\item the development of a linear equisatisfiable translation of
  \TDLLite\ KBs into \LTL formulas, with formal proofs for the
  strategy that removes past operators over the integers;
\item an experimental analysis measuring the runtime, and the size of
  the translation based on randomly generated \TDLLite\ KBs, and an
  experimental analysis of toy scenarios exploiting the modelling
  capabilities of the language.  We compare the performance of
  different \LTL solvers on the various test cases we consider. For
  the randomly generated ontologies, we present benchmarks for the
  following cases: $(i)$ \TDLLite\ TBoxes; and $(ii)$ \TDLLiteN\
  TBoxes. As for the toy scenario, we evaluate the perfomances with
  ABoxes of increasing sizes.
\end{enumerate}


In the following, we present our running examples to provide an
intuitive overview on the expressive power of \TDLLite. These examples
are also part of our experimental tests.
\begin{example} \label{ex:1} Let's consider first the simple case of a
  person who might be minor or adult. This can be modelled by the
  following \TDLLite\ TBox:
\begin{align*}
 \T=\{{\sf Adult} \sqsubseteq {\sf Person}, {\sf Minor} \sqsubseteq {\sf Person}, {\sf Minor} \sqcap {\sf Adult} \sqsubseteq \bot, {\sf Adult} \sqsubseteq \B_F {\sf Adult}\}
\end{align*}  
In words, $\T$ states that minors and adults are persons, but they are
disjoint. Moreover, adult is persistent in the future---expressed by
the temporal operator \emph{always in the future} $\B_F$.  Now, assume
one wants to check the consistency of the following ABox,
reporting on the status of the person {\it John} at different
timestamps, w.r.t. the TBox \T:
\begin{align*}
 \A=\{{\sf Person} ({\sf John},0),{\sf Minor}({\sf John},0),{\sf Adult}({\sf John},1),{\sf Minor}({\sf John},2),{\sf Adult}({\sf John},3)\}
\end{align*}
According to the persistence of being adult together with the fact
that adult is disjoint from minor, the ABox $\A$ is inconsistent
w.r.t. \T.  \hfill {\mbox{$\triangleleft$}}
\end{example}

In the following example we show another critical case of
inconsistency due to the interaction between cardinalities and
\emph{global roles}, i.e., binary relations whose instances are
time-invariant.

\begin{example}
\label{ex:2}
Consider the following knowledge base, $\K=(\T,\A)$, where the TBox
assumes that each person has a {\sf Name} being a global functional
role (role names\nb{A: added} are distinguished being either
\emph{global} or \emph{local} in the signature of the logic, see next
Section):
%
\begin{align*}
\T &=\{{\sf Person} \sqsubseteq \ge 1~{\sf Name}, {\sf Person}
     \sqsubseteq \lnot \ge 2~{\sf Name}\}\\
 \A &=\{{\sf Person}(p_1,0), {\sf Name}(p_1,{\sf\textit{Kennedy}},0), {\sf Name}(p_1,{\sf\textit{Marc}},1)\}  
\end{align*}
In words, $\T$ states that each person has a single name which, in
turns, is a global role. On the other hand, while at each point in
time, $p_1$ has exactly one name, $p_1$ has different names at
different points in time. Thus, $\A$ violates the fact that name is
both functional and global, and $\K$ is then inconsistent.  \hfill
{\mbox{$\triangleleft$}}
%
%
\end{example}

This paper is organised as follows. Section~\ref{sec-tdl} introduces
the temporal DL \TDLLite. Section~\ref{sec-translation} describes the
equisatisfiable encoding from \TDLLite to \LTL.
Section~\ref{sec-impl} describes the architecture of the tool we are
proposing. Section~\ref{sec-eval} illustrates the experimental setting
and reports on the performances obtained by using different solvers to
reason over \TDLLite KBs. We report our concluding remarks in
Section~\ref{sec:conclusion}.


%% file: 2-preliminaries.tex
\section{Temporal Description Logic}
\label{sec-tdl}

We briefly introduce the syntax and the semantics of the temporal
description logic \TDLLite. In this paper by \TDLLite we denote the
logic $\TdxDLbn$, the fragment of $\TuDLite{}{bool}$, introduced
in~\cite{ArtEtAl3}, that allows the past and future temporal operators
$\Box,\Diamond, \Next$. We notice that 
while the complexity of \TuDLite{}{bool}, which allows for both \Umc
and \Smc, is the same as $\TdxDLbn$, \Umc and \Smc are barely used in
conceptual modelling~\cite{ArtEtAl3,AKRZ:ER10}. We consider in this
paper the non-strict semantics of the diamond and box operators (as
usual, strictness can be expressed using the next operator, e.g.,
strict diamond in the future is $\Rnext\Rdiamond C$).
%
Let $\NC, \NI$ be countable sets of \emph{concept} and
\emph{individual names}, respectively, and let $\NGl$ and $\NL$ be
countable and disjoint sets of \emph{global} and \emph{local role
  names}, respectively.  The union $\NGl \cup \NL$ is the set $\NR$ of
\emph{role names}.  $\TDLLite$ \emph{roles} $R$, \emph{basic concepts}
$B$, and \emph{(temporal) concepts} $C$ are given by the following
grammar:
\begin{align*}
R & ::=  L \mid L^{-} \mid G \mid G^{-}, & B & ::=  \bot \mid A \mid
                                               \, \geq q R, &
 \\
  C & ::= B \mid \lnot C \mid C_{1} \sqcap C_{2}\mid \D_F C \mid \D_P C
      \mid \Next_F C \mid  \Next_P C,
\end{align*}
where $L \in \NL$, $G \in \NGl$, $A \in \NC$, and $q \in \Nbl, q > 0$
(given in binary).

We use standard abbreviations for concepts: $\top := \lnot \bot$,
$(C_{1} \sqcup C_{2}) := \lnot(\lnot C_{1} \sqcap \lnot C_{2})$,
$\exists R := \geq 1 R$,\nb{A: added for REV1}
$\Next^{n + 1} C := \Next_F \Next_F^{n} C$, with $n \geq 0$ (we set
$\Next^{0} C := C$), $\Next^{n - 1} C := \Next_P \Next_P^{n} C$, with
$n < 0$, $\B_F C := \neg\D_F\neg C$, $\B_P C := \neg\D_P\neg C$,
$\SVbox C = \Rbox\Lbox C$, and $\SVdiamond C = \Rdiamond\Ldiamond
C$. 

A \TDLLite \emph{axiom} is either a \emph{concept inclusion (CI)} of
the form $C_{1} \sqs C_{2}$, a \emph{concept assertion} of the form
$\Next^{n}A(a)$ or $\Next^{n}\neg A(a)$, or a \emph{role assertion} of
the form $\Next^{n}R(a,b)$ or $\Next^{n}\neg R(a,b)$, where
$C_{1}, C_{2}$ are \TDLLite concepts, $A\in\NC$, 
$R\in\NR$, $a,b\in\NI$, and\nb{A: added} $n\in \mathbb{Z}$.
A \emph{TBox} is a set of CIs and an \emph{ABox} is a set of (concept
and role) assertions.  A \TDLLite \emph{knowledge base}, $\K$, is a
pair $\K = (\T,\A)$, where $\T$ is a TBox and $\A$ is an ABox.
%
%

A \TDLLite \emph{interpretation} is a structure
$\Mmf=(\Delta^{\Mmf}, (\I_{n})_{n \in \mathbb{Z}})$, where each
$\I_{n}$ is a classical DL interpretation with non-empty domain
$\Delta^\Mmf$ (or simply $\Delta$).  We have that
$A^{\I_{n}} \subseteq \Delta^\Mmf$ and
$S^{\I_{n}} \subseteq \Delta^\Mmf \times \Delta^\Mmf$, for all
$A \in \NC$ and $S \in \NR$. In particular, for all
$G \in \textsf{N}_{\textsf{G}}$ and $i,j \in \Zbb$,
$G^{\I_{i}} = G^{\I_{j}}$ (denoted simply by $G^{\I}$).
Moreover, $a^{\I_{i}}=a^{\I_j}\in \Delta^\Mmf$ for all $a\in\NI$ and
$i,j \in \Zbb$, i.e., constants are \emph{rigid designators} (with
fixed interpretation, denoted simply by $a^{\I}$). The interpretation
of individuals further respects the unique name assumption UNA, i.e., $a^{\I} \neq b^{\I}$ if
$a\neq b$.
The stipulation that all  time points share
the same domain $\Delta^\Mmf$ is called the \emph{constant domain
assumption} (meaning that objects are not created nor destroyed over
time).
%
The interpretation of roles and concepts at instant $n \in \Zbb$ is
defined as follows (where $S \in \NR$):
\begin{gather*}
(S^{-})^{\I_{n}} = \{ (d', d)\in  \Delta^\Mmf \times \Delta^\Mmf \mid (d, d') \in S^{\I_{n}} \}, \qquad
\bot^{\I_{n}} = \eset, \\
(\geq q R)^{\I_{n}} = \{d \in \Delta^\Mmf \mid \sharp\{ d' \in \Delta^\Mmf \mid (d, d') \in  R^{\I_{n}} \} \geq q \}, \\
(\neg C)^{\I_{n}} = \Delta^\Mmf \setminus C^{\I_{n}}, \qquad 
(C_{1} \sqcap C_{2})^{\I_{n}} = C_{1}^{\I_{n}} \cap C_{2}^{\I_{n}}, \\
(\Next_F C)^{\I_{n}} = \{ d \in \Delta^{\Mmf} \mid \ d \in C^{\I_{n+1}}\},\quad
(\Next_P C)^{\I_{n}} = \{ d \in \Delta^{\Mmf} \mid \ d \in C^{\I_{n-1}}\}, \\
(\D_F C)^{\I_{n}} = \{ d \in \Delta^{\Mmf} \mid \exists m \in \Zbb, m \geq n \colon d \in C^{\I_{m}} \}, \\
(\D_P C)^{\I_{n}} = \{ d \in \Delta^{\Mmf} \mid \exists m \in \Zbb, m \leq n \colon d \in C^{\I_{m}} \}.
\end{gather*}

We say that a concept $C$ is \emph{satisfied in \Mmf} if there is
$n\in\Zbb$ such that $C^{\I_{n}} \neq \eset$.  The \emph{satisfaction
  of an axiom in $\Mmf$} is defined as follows:
\[
\begin{array}{llllll}
\Mmf \models C_1\sqsubseteq C_2  &\text{iff}& C_1^{\I_{n}} \subseteq C_2^{\I_{n}} \text{ for all } n\in\Zbb,&&&\\
\Mmf  \models \Next^n A(a)  &\text{iff}& a^{\I} \in A^{\I_{n}}, &
\Mmf  \models \Next^n \neg A(a)  &\text{iff}& a^{\I} \not\in A^{\I_{n}},\\
\Mmf \models \Next^n  R(a,b)  &\text{iff}& (a^{\I},b^{\I}) \in R^{\I_{n}}, &
\Mmf \models \Next^n \neg R(a,b)  &\text{iff}& (a^{\I},b^{\I}) \not\in R^{\I_{n}}.
 \end{array}
\]
         %
CIs are interpreted globally while assertions are
interpreted relative to the initial time point, $0$.  A
KB $\Kmc = (\Tmc, \Amc)$ is \emph{satisfiable},  written
$\Mmf \models \K$, if all axioms in $\T$ and \Amc are satisfied in some $\Mmf$. 

%% file: 3-translation.tex
\section{Reduction to \LTL}
\label{sec-translation}

This section contains the reduction of a \TDLLite KB, \K, into an \LTL
formula. In the following, we distinguish \PTL, using both past and
future operators interpreted\nb{A: added} over $\mathbb{Z}$, from
\LTL, using only future operators interpreted\nb{A: added} over
$\mathbb{N}$.  We first report the equisatisfiable reduction presented
in~\cite{ArtEtAl3}, where \K\ is first reduced to a first-order
temporal formula with one free variable, \QTLO, and then to an \PTL
formula.  We then present an equisatisfiable translation of the \PTL
formula into \LTL.
%
We consider also the simpler case where there are no
temporal past operators, and axioms
(including ABox assertions) are interpreted over $\Nbb$ (with the
obvious semantics). We denote this language by \TDLLiteN. 

\subsection{Reduction to \QTLO}
\label{QTL-reduction}

To define the translation of a \TDLLite KB into
\QTLO---the fragment of first-order temporal formulas with one free variable---
we first define the translation of concepts (see~\cite[Sect.4.2]{ArtEtAl3}).
%
Given a \TDLLite concept $C$, we inductively define the \QTLO-formula
$C^*(x)$ as:
\begin{align*}
  A^* & = A(x), & \bot^* & =\bot,& (\mathop{\geq q} R)^* & = E_qR(x), \\ %
(\mathbb{O} C)^* & = \mathbb{O} C^*, &
 (C_1 \sqcap C_2)^* &= C_1^* \land  C_2^*, & (\neg C)^* &= \neg C^*.
 \end{align*}
where $\mathbb{O}\in\{\Rnext, \Lnext, \Rdiamond, \Ldiamond, \Rbox,
\Lbox\}$, and\nb{A: added} $E_qR(x)$ unary predicates capturing the
at-least cardinalities for roles.
%
Now, the translation $\T^\dagger $ of a TBox $\T$ is the conjunction
of:\nb{A: corrected~\eqref{eq:role:existence:2} as suggested by REV2}
 \begin{align}
 \label{eq:tbox}&\bigwedge_{C_1 \sqsubseteq C_2\in\T}
\hspace*{-1em} \SVbox\forall x\,\bigl(C_1^*(x) \to C_2^*(x)\bigr),\\
  \label{eq:role:saturation}&\bigwedge_{R\in\role}\hspace*{1em}
  \bigwedge_{q,q'\in \QT \text{ with } q' > q}\hspace*{-1em}\SVbox\ \forall
  x\,\bigl((\mathop{\geq q'}R)^*(x) \to (\mathop{\geq q}R)^*(x)\bigr),\\
  \label{eq:role:global:relation}&
  \bigwedge_{R \in\role \text{ is global}} \hspace*{1em}\bigwedge_{q\in
    \QT}\hspace*{-0.4em} \SVbox\ \forall x\,\bigl((\mathop{\geq q}R)^*(x)
  \rightarrow \SVbox\,(\mathop{\geq q}R)^*(x)\bigr),\\%
  \label{eq:role:existence:2} & \bigwedge_{R\in\role} \SVbox\, \forall x\,
  \bigl((\exists R)^*(x) \ \rightarrow \ \exists
  x\,(\exists \inv(R))^*(x)\bigr),
\end{align}%
where $\role$ is the set of (global and local) role names occurring in
$\K$ and their inverses, $\QT$\label{def:QK} is the union of $\{1\}$
and the set of all numbers occurring in $\T$,
and $\inv(R)$ is the inverse of $R$ (that is, $\inv(R) = R^-$ and
$\inv(R^-) = R$, for all $R\in\NR$).

It remains to explain how an ABox $\A$ is translated. For each
$n\in \Z$ and each role $R$, we define:
$\A_n^R:=\{ R(a,b) \mid \nxt^m R(a,b)\in \A \text{ for some
  }m \in \Z \bigr\}$, if $R\in \NGl$, and
  $\A_n^R:=\{ R(a,b) \mid \nxt^n R(a,b) \in \A \bigr\}$, if  $R\in \NL$.
  Let $S \in \NR$, we may sssume w.l.o.g. that $\nxt^n S(a,b)\in\A$
  implies $\nxt^n S^-(b,a)\in\A$.\label{def:invABox} The translation
  $\A^\dagger$ of \A \ is
\begin{equation}
   \label{eq:abox}
  \A^\dagger = \hspace*{-0.5em}
\ \ \Phi_\Amc \ \ \land\hspace*{-0.2em}
  \bigwedge_{{\scriptscriptstyle\bigcirc}^n R(a,b)\in \A}\hspace*{-1.5em} \nxt^n
  (\mathop{\geq q^{R,n}_{\A(a)}} R)^*(a)\hspace*{1em}\land
  \bigwedge_{\begin{subarray}{c}{\scriptscriptstyle\bigcirc}^n \neg
      S(a,b)\in \A\\S(a,b)\in \A_n^S \end{subarray}}\hspace*{-1.5em} \bot,
\end{equation}
where $\Phi_\Amc$ is the conjunction of all concept assertions in \Amc
and $\smash{q^{R,n}_{\A(a)}}$ is the maximun between $1$ and the
number of distinct $b$ such that $R(a,b)\in\A_n^R$.  The last two
conjuncts of $\A^\dagger$ are important to ensure that (a) the number
of role successors of individuals is represented using the predicates
$E_qR$\nb{A: corrected typo as REV2} that appear in the translation
for concepts and TBoxes, and that (b) the semantics of global roles is
preserved by the translation.  The translation of $\Kmc=(\T,\A)$ into
$\Kmc^\dagger=\T^\dagger\wedge\A^\dagger$ is correct and can be
computed in polynomial time~\cite{ArtEtAl3}.
%
\begin{theorem}\label{lem:qtli-equisat}~\cite{ArtEtAl3}
  A \TDLLite KB $\Kmc$ is satisfiable iff the \QTLO-formula
  $\Kmc^\dagger$ is satisfiable. Moreover, $\Kmc^\dagger$
  can be constructed in polynomial time w.r.t. the size of \Kmc.
\end{theorem}

The translation of a \TDLLiteN KB 
is defined in the same way as for a \TDLLite KB, except that every
outer $\SVbox$ is replaced by $\Box_F$, while to deal with global
roles~\eqref{eq:role:global:relation} becomes: 
 \begin{align*}
 \label{eq:role:global:relation:N}&
  \bigwedge_{R \in\role \text{ is global}} \hspace*{1em}\bigwedge_{q\in
    \QT}\hspace*{-0.4em} \Box_F\ \forall x\,\bigl(\Diamond_F(\mathop{\geq q}R)^*(x)
  \rightarrow \Box_F\,(\mathop{\geq q}R)^*(x)\bigr),
\end{align*}%

\begin{corollary}\label{lem:qtli-equisat} 
  A \TDLLiteN KB $\Kmc$ is satisfiable iff the \QTLO-formula
  $\Kmc^\dagger$ is satisfiable. Moreover, $\Kmc^\dagger$ can be
  constructed in polynomial time w.r.t. the size of \Kmc.
\end{corollary}

\subsection{Reduction to \PTL}
\label{PTL-reduction}

Here we present the conversion of the \QTLO translation in
Subsection~\ref{QTL-reduction} into an equisatisfiable
\PTL-formula~\cite[Sect.4.3]{ArtEtAl3}.  As usual, this is done by
grounding the formula.  Let $\K^\dagger$ be the \QTLO translation of a
\TDLLite KB $\Kmc=(\Tmc,\Amc)$.  Assume w.l.o.g. that $\K^\dagger$
is of the form:
\begin{equation*}
 \ \ \SVbox\forall x\,\varphi(x) \  \land \bigwedge_{R\in\role} \SVbox\, \forall x\, \bigl( (\exists R)^*(x) \rightarrow \exists x\,(\exists \inv(R))^*(x)\bigr)\land \ \Amc^\dagger
\end{equation*}
where $\varphi(x)$ is a quantifier-free \QTLO 
formula
with a single variable $x$ and only unary predicates. 
Now, consider the formula $\K^{\dagger'}$ 
\begin{equation*}
\ \ \SVbox\forall x\,\varphi(x) \  \land \bigwedge_{R\in\role} \SVbox\forall x\,\bigl((\exists R)^*(x)
 \to \SVbox p_R\bigr)
 \land
   \bigl(p_{\inv(R)} \to (\exists R)^*(d_{R}) \bigr)\land \ \Amc^\dagger
\end{equation*}
where, for each $R\in\role$, $d_R$ is a fresh constant and $p_R$ is a
fresh propositional variable. This formula does not have existential
quantifiers. We define $\K^{\ddagger}$ as the result of grounding
$\K^{\dagger'}$ using all constants in the formula so that the
universal quantifiers can also be removed.
\begin{theorem}\label{lem:qtli-equisat2}~\cite{ArtEtAl3}
  The \PTL translation $\Kmc^\ddagger$ of a \TDLLite KB $\Kmc$ is
  satisfiable iff \Kmc is satisfiable. Morever, $\Kmc^\ddagger$ can be
  constructed from $\Kmc^\dagger$ in logarithmic space w.r.t. the size
  of \Kmc.
\end{theorem}

We now consider the \TDLLiteN case.  Since there are no past
operators, we make the translation directly into \LTL (it is also a
translation into \PTL). 
Assume $\Kmc^\dagger$ is the translation of a \TDLLiteN KB into
\QTLO. The formula $\Kmc^{\dagger'}$ is the same as
$\Kmc^{\dagger'}$ in the \TDLLite case, except that every
$\SVbox$ is replaced by $\Box_F$ and
the  conjuncts ranging over $\role$ are  of the form: 
%
%
\begin{equation*}
\Box_F\forall x\,\bigl[\Diamond_F(\exists R)^*(x)
 \to \Box_F p_R\bigr]
 \land
   \bigl[p_{\inv(R)} \to (\exists R)^*(d_{R}) \bigr]
 \end{equation*}
 The translation $\Kmc^\ddagger$ is now obtained by grounding $\Kmc^{\dagger'}$.
\begin{corollary}\label{lem:qtli-equisat2} 
  The \QTLO translation $\Kmc^\dagger$ of a \TDLLiteN KB $\Kmc$ is
  satisfiable iff the \LTL-formula $\Kmc^\ddagger$ is
  satisfiable. Also, $\Kmc^\ddagger$ can be constructed in polynomial
  time w.r.t. the size of \Kmc.
\end{corollary}



\subsection{Reduction 
  from \PTL to  \LTL}
\label{Future-reduction} 

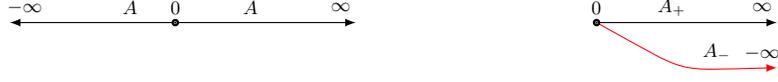
\begin{figure}[t]
\begin{center}	
		\input{bending.tikz}
\end{center}
\caption{\Zbb timeline (left) and bending \Zbb\ over \Nbb (right)}
\label{fig:bending}
\end{figure}

%

Due\nb{A: small changes} to the inability of various off-the-shelf
reasoners to deal with past operators, we perform a further
translation of the $\PTL$-formula $\Kmc^\ddagger$, as defined in the
previous section and containing both future and past temporal
modalities, into a pure-future formula thus expressed in \LTL (i.e., a
formula using just future temporal modalities). In\nb{A: added some
  remarks and citations} this respect, Gabbay~\cite{Gab80} showed that
past temporal modalities do not add expressive power providing also an
algorithm~\cite{Gab89} for translating formulas with past into
pure-future formulas, preserving \emph{formula equivalence}. While
Gabbay's algorithm produces pure-future formulas of size
non-elementary in the size of the input formula, recently
Markis~\cite{Markey03} has presented an algorithm that produces
equivalent pure-future formulas with an exponential blow-up.
%
%
On the other hand,\nb{A: small changes} in case we want to maintain
satisfiability, a linear in size translation, inspired by ideas from
normal forms in propositional logic~\cite{Tseytin66}, that removes
past operators and preserves satisfiability when formulas are
interpreted over the natural numbers has been presented
in~\cite{DBLP:conf/lpar/GiganteMR17}.
%

In our case, \nb{A: added} we are interested in checking satisfiability
of formulas with both future and past operators interpreted over
$\Zbb$.  Thus, in the following we present a linear in size
translation removing past operators preserving satisfiability when
formulas are interpreted over the integers.  Let \subk be the set of
all subformulas of $\K^{\ddagger}$, the translation
$\K^{\ddagger_{\Nbb}}$ of $\K^{\ddagger}$ will be defined over the
alphabet $\Sigma_{\K^{\ddagger_{\Nbb}}}$ containing a pair of
propositional variables $A_+, A_-$ for each propositional variable $A$
in \subk together with a pair of propositional variables
$A_+^{\mathbb{O}\psi}, A_-^{\mathbb{O}\psi}$ for every
temporal 
formula $\mathbb{O}\psi \in \textit{sub}(\K^{\ddagger})$, where
$\mathbb{O}$ stands for any of the following temporal operators:
$\Rnext, \Lnext, \Rdiamond, \Ldiamond, \Rbox, \Lbox$.
%
The main intuition is that in a model over \Nbb 
the propositional variable $A_+$ at the moment of time $n \geq 0$ has
the same truth value as the propositional variable $A$ in a model over
\Zbb, whereas the propositional variable $A_-$ has the same truth
value as $A$ at the moment of time $-n \leq 0$, with the truth value
of $A_+, A_-$ coinciding at time $0$ (similarly for
$A_+^{\mathbb{O}\psi}$ and $A_-^{\mathbb{O}\psi}$).
Intuitively, we are `bending' the negative part of the time
line obtaining two parallel time lines. The first (second,
respectively) will represent the future (past) and will only be used
to evaluate propositional variables $A_+$ and $A^{\mathbb{O}\psi}_+$
($A_-$ and $A^{\mathbb{O}\psi}_-$) as shown in Figure~\ref{fig:bending}. 
For each $\xi \in \subk$, we define the translations
$\overline{\xi}_+$ and $\overline{\xi}_-$ to formulas of
propositional logic over the newly defined alphabet,
$\Sigma_{\K^{\ddagger_{\Nbb}}}$, as follows:
%
\begin{align*}
\overline{\xi}_\ast=
\begin{cases}
  A_\ast,&\text{ if }\xi=A,\\
  \neg \overline{\psi}_\ast, &\text{ if }\xi=\neg \psi,\\
  \overline{\psi}_\ast \wedge \overline{\phi}_\ast, &\text{ if
  }\xi=\psi \wedge \phi, 
  \\
  A_\ast^{\mathbb{O}\psi},&\text{ if }\xi = \mathbb{O}\psi
\end{cases}
\end{align*}%
where either all $\ast$ is $+$ or $-$.
Finally, we define the \LTL\ translation $\K^{\ddagger_{\Nbb}}$ of
$\K^{\ddagger}$ as (recall that $\Box_{P/F}$ is a syntactic sugar,
omitted from the translation for simplicity):
{\small
\begin{align*}
&\K^{\ddagger_{\Nbb}} :=  \overline{\K^{\ddagger}}_+ \land \bigwedge_{\overline{\xi}_+ \in \Sigma_{\K^{\ddagger_{\Nbb}}}} (\overline{\xi}_+ \leftrightarrow \overline{\xi}_-) \ \wedge
\\
& \Box_F \Bigg[ \bigwedge_{\Next_F \psi \ \in \ \subk} \Big( \big(\Next_F \ \overline{\Next_F \psi}_- \leftrightarrow \overline{\psi}_-\big) \ \ \wedge \ \big(\overline{\Next_F \psi}_+ \leftrightarrow \Next_F \overline{\psi}_+ \big)\Big) \ \wedge 
\\
&\bigwedge_{\Next_P \psi \ \in \ \subk} \Big( \big(\Next_F \ \overline{\Next_P \psi}_+ \leftrightarrow \overline{\psi}_+\big) \ \wedge \ \big(\overline{\Next_P \psi}_- \leftrightarrow \Next_F \overline{\psi}_- \big)\Big) \ \wedge \\
&\bigwedge_{\Diamond_F \psi \ \in \ \subk} \Big( \big( \Next_F
\overline{\Diamond_F \psi}_- \leftrightarrow (\overline{\Diamond_F
  \psi}_- \ \vee \ \Next_F \overline{\psi}_- )\big) \ \wedge \
\big(\overline{\Diamond_F \psi}_+ \leftrightarrow \Diamond_F
\overline{\psi}_+\big)\Big) \ \wedge\\
%
 &\bigwedge_{\Diamond_P \psi \ \in \ \subk} \Big( \big( \Next_F \overline{\Diamond_P \psi}_+ \leftrightarrow (\overline{\Diamond_P \psi}_+ \ \vee \ \Next_F \overline{\psi}_+) \big) \ \wedge \ \big(\overline{\Diamond_P \psi}_- \leftrightarrow \Diamond_F \overline{\psi}_- \big)\Big)
\Bigg]
\end{align*}%
}
The size of $\K^{\ddagger_{\Nbb}}$ is linear in the size of the
original formula $\K^{\ddagger}$ at the cost of extending the alphabet
with new propositional variables. 
The extension of the alphabet is unavoidable as it is well-known that
\PTL\ is exponentially more succinct than an equivalent \LTL\
formula~\cite{Markey03}. By an inductive argument we can prove the
following.
\begin{theorem}\label{lem:qtli-equisat}
  $\K^{\ddagger}$ is satisfiable iff the pure future translation
  $\K^{\ddagger_{\Nbb}}$ is satisfiable.
  \begin{proof}\upshape
    ($\Rightarrow$) In the following we are using the notation
    $\Mmf=(\Delta^{\Mmf}, (\I_{n})_{n \in \Nbb{}})$ as an \LTL
    interpretation with the obvious meaning. Let
    $\Mmf, 0\models \K^{\ddagger}$. We then define an interpretation
    $\Mmf'=(\Delta^{\Mmf}, (\I'_{n})_{n \in \Nbb{}})$ 
    over the alphabet $\Sigma_{\K^{\ddagger_{\Nbb}}}$, as
    follows, for $n\geq 0$:
    \begin{align*}
      A_+^{\I'_{n}} = A^{\I_{n}},\quad A_-^{\I'_{n}} =
      A^{\I_{-n}},\quad (A_+^{\mathbb{O}\psi})^{\I'_{n}} =
      (\mathbb{O}\psi)^{\I_{n}}, \quad  (A_-^{\mathbb{O}\psi})^{\I'_{n}} =
      (\mathbb{O}\psi)^{\I_{-n}}.
    \end{align*}
    By an easy induction argument and the definition of $\Mmf'$ the
    following holds:
    \begin{lemma}\label{lemma-sat}
      For any $\psi \in \subk$, for any $\Mmf$, and any $n\in\Nbb$ the following
      holds:
      \begin{align}
        \Mmf, n\models \psi &\quad\textit{iff}\quad \Mmf', n\models \overline{\psi}_+\\
        \Mmf, -n\models \psi &\quad\textit{iff}\quad \Mmf', n\models \overline{\psi}_-
        \end{align}
      \end{lemma}
      From the above lemma it follows that
      $\Mmf',0\models \overline{K^{\ddagger}}_+$, while the fact that
      $\Mmf',0\models \bigwedge_{\overline{\xi}_+ \in
        \Sigma_{\K^{\ddagger_{\Nbb}}}} (\overline{\xi}_+
      \leftrightarrow \overline{\xi}_-)$ is a direct consequence of
      the definition of $\Mmf'$.
    
      It remains to be shown that for any temporal subformula in
      $\subk$ the corresponding conjunct in $\K^{\ddagger_{\Nbb}}$ is
      satisfiable in $\Mmf'$. We show the case where
      $\Diamond_F\psi\in\subk$. We start by showing that, for all
      $n\geq 0$,
      $\Mmf',n\models (\overline{\Diamond_F \psi}_+ \leftrightarrow
      \Diamond_F \overline{\psi}_+\big)$. By $\Mmf'$ definition,
      $\Mmf',n\models \overline{\Diamond_F \psi}_+$ iff
      $\Mmf,n\models \Diamond_F \psi$, iff, $\exists m\geq n$
      s.t. $\Mmf, m \models \psi$, iff, by Lemma~\ref{lemma-sat},
      $\Mmf',m\models \overline{\psi}_+$, iff,
      $\Mmf',n\models \Diamond_F \overline{\psi}_+$. We now show that,
      for all $n\geq 0$,
      $\Mmf',n\models \big( \Next_F\overline{\Diamond_F \psi}_-
      \leftrightarrow \overline{\Diamond_F\psi}_- \ \vee \ \Next_F\overline{\psi}_- \big)$.
      $\Mmf',n\models \Next_F \overline{\Diamond_F \psi}_-$ iff
      $\Mmf',n+1\models\overline{\Diamond_F \psi}_-$, iff, by
      definition of $\Mmf'$, $\Mmf,-n-1\models \Diamond_F \psi$,
      iff, either $\Mmf,-n-1\models \psi$ or
      $\Mmf,-n\models \Diamond_F \psi$ iff, by Lemma~\ref{lemma-sat},
      either $\Mmf',n+1\models \overline{\psi}_-$ or
      $\Mmf',n\models \overline{\Diamond_F \psi}_-$ iff
      $\Mmf',n\models \Next_F\overline{\psi}_-$ or
      $\Mmf',n\models \overline{\Diamond_F \psi}_-$. The other
      temporal subformulas can be treated in a similar way.

    ($\Leftarrow$) We show the following lemma from which this
    direction easily follows.
    \begin{lemma}
      Let $\varphi$ be a \PTL formula, $\varphi^{\Nbb}$ its
      pure-future translation and $\Mmf'$ a model, i.e.,
      $\Mmf',0\models \varphi^{\Nbb}$, then there exists an
      interpretation $\Mmf$ s.t., for every
      $\psi\in\subphi$ and $n\in\Zbb$, the following holds
      $$
      \Mmf,n \models \psi \textit{ iff }
      \begin{cases}
        \Mmf',n \models \overline{\psi}_+, \textit{ if } n\geq 0\\
        \Mmf',-n \models \overline{\psi}_-, \textit{ if } n< 0
      \end{cases}
      $$
      \begin{proof}\upshape
        We define the interpretation $\Mmf$ as follows, for each
        propositional atom $A \in \subphi$:
      $$
      \Mmf,n \models A \textit{ iff }
      \begin{cases}
        \Mmf',n \models \overline{A}_+, \textit{ if } n\geq 0\\
        \Mmf',-n \models \overline{A}_-, \textit{ if } n< 0
      \end{cases}
      $$
      The proof is by structural induction. The base case holds by
      definition of $\Mmf$.\\
      $\psi = \psi_1\land \psi_2$. Let $n\geq 0$, $\Mmf,n \models \psi$, iff,
      $\Mmf,n \models \psi_1$ and $\Mmf,n \models \psi_2$, iff, by
      induction, $\Mmf',n \models \overline{\psi_1}_+$ and
      $\Mmf',n \models \overline{\psi_2}_+$, iff,
      $\Mmf',n \models \overline{\psi_1\land\psi_2}_+$. Similarly for the case $n<0$.\\
      In a similar way we can prove the other boolean connectives. We
      thus proceed with the temporal subformulas.\\
      $\psi = \Rdiamond \alpha$. We use
      the following facts (that hold by the assumption that
      $\Mmf',0\models \varphi^{\Nbb}$): for all $\ell\in\Nbb$
      \begin{align}
        \label{eqn-dia-1}
        \Mmf',\ell\models &\big( \Next_F \overline{\Diamond_F \alpha}_- \leftrightarrow
                            \overline{\Diamond_F \alpha}_- \ \vee \ \Next_F
                            \overline{\alpha}_- \big)\\
        \label{eqn-dia-2}
        \Mmf',\ell\models &\big(\overline{\Diamond_F \alpha}_+
                            \leftrightarrow \Diamond_F\overline{\alpha}_+\big)
      \end{align}
      Now, $\Mmf,n \models \psi$ iff (we
      consider the nearest $m$ where $\alpha$ holds)
      \begin{align}
        \label{diamond}
        \exists m\geq n \text{ such that } \Mmf,m \models \alpha \text{
        and, for any } n\leq m'< m,~ \Mmf,m' \not\models \alpha.
      \end{align}
      We distinguish the case $n\geq 0$ from $n<0$. Let's first
      consider $n\geq 0$. Then, $m\geq 0$ and~\eqref{diamond} holds
      iff, by induction, $\Mmf',m \models \overline{\alpha}_+$, and
      $\Mmf',m' \not\models \overline{\alpha}_+$, iff,
      $\Mmf',n \models \Rdiamond\overline{\alpha}_+$, iff,
      by~\eqref{eqn-dia-2},
      $\Mmf',n \models \overline{\Rdiamond\alpha}_+$. Let's now consider
      the case $n<0$. Here we distinguish the two directions of the
      iff, and we further distinguish whether $m\geq 0$ or
      $m<0$.
      Let's first consider the case $m<0$, then, by~\eqref{diamond}
      and induction, $\Mmf,n \models \psi$ iff
      \begin{align}
        \label{diamond1}
        \Mmf',-m \models \overline{\alpha}_-, \text{ and }
        \Mmf',-m' \not\models \overline{\alpha}_-
      \end{align}
      If~\eqref{diamond1} holds, then,
      $\Mmf',-m-1 \models \Next_F\overline{\alpha}_-$, and
      by~\eqref{eqn-dia-1},
      $\Mmf',-m-1 \models \Next_F\overline{{\Diamond_F \alpha}}_-$,
      i.e., $\Mmf',-m \models \overline{{\Diamond_F \alpha}}_-$. Now,
      either $m=n$ and we are done, or again by~\eqref{eqn-dia-1},
      $\Mmf',-n \models\overline{{\Diamond_F \alpha}}_-$. Let now be
      $m\geq 0$, then, by~\eqref{diamond} and induction,
      $\Mmf,n \models \Rdiamond\alpha$ iff
      \begin{multline}
        \label{diamond2}
        \Mmf',m \models \overline{\alpha}_+, \text{ and } 
        \Mmf',m' \not\models \overline{\alpha}_+, \text{ for any }
        0\leq m'<m, \text{ and }\\ \Mmf',-m'' \not\models \overline{\alpha}_-, \text{ for any }
        n\leq m''<0
      \end{multline}
      Then, $\Mmf',0 \models \Rdiamond\overline{\alpha}_+$, and
      by~\eqref{eqn-dia-2},
      $\Mmf',0 \models \overline{\Rdiamond\alpha}_+$, and by the
      second conjunct in $\varphi^\Nbb$,
      $\Mmf',0 \models \overline{\Rdiamond\alpha}_-$, and
      by~\eqref{eqn-dia-1},
      $\Mmf',-n \models\overline{{\Diamond_F \alpha}}_-$. It remains
      to be proved the viceversa, for the case $n<0$, i.e., if
      $\Mmf',-n \models\overline{{\Diamond_F \alpha}}_-$ then
      $\Mmf,n \models \Diamond_F \alpha$. It's enough to show that
      either~\eqref{diamond1} or~\eqref{diamond2} hold. Indeed, let
      $\Mmf',-n \models\overline{{\Diamond_F \alpha}}_-$, then,
      $\Mmf',-n-1 \models\Next_F\overline{{\Diamond_F \alpha}}_-$, and
      by~\eqref{eqn-dia-1}, either there is $m$ such that $n\leq m <0$
      and~\eqref{diamond1} holds, or
      $\Mmf',0 \models\overline{{\Diamond_F \alpha}}_-$, and by the
      second conjunct in $\varphi^\Nbb$,
      $\Mmf',0 \models \overline{\Rdiamond\alpha}_+$, and
      by~\eqref{eqn-dia-2},
      $\Mmf',0 \models \Rdiamond
      \overline{\alpha}_+$. Thus,~\eqref{diamond2} holds and we are
      done.\\
      The other cases are similar and left to the reader.
    \end{proof}
  \end{lemma}
\end{proof}
\end{theorem}

%% file: bending.tikz
\begin{tikzpicture}[scale=0.8, transform shape]
	\begin{pgfonlayer}{nodelayer}
		\node [style=newstyle1] (0) at (-3, 1.25) {};
		\node [style=none] (1) at (0, 1.25) {};
		\node [style=none] (2) at (-5.75, 1.25) {};
		\node [style=none] (3) at (7, 1.25) {};
		\node [style=newstyle1] (4) at (4, 1.25) {};
		\node [style=none] (5) at (7, 0.5) {};
		\node [style=none] (6) at (4, 1.5) {$0$};
		\node [style=none] (7) at (-0.25, 1.5) {$\infty$};
		\node [style=none] (8) at (-5.5, 1.5) {$-\infty$};
		\node [style=none] (9) at (6.75, 1.5) {$\infty$};
		\node [style=none] (10) at (6.75, 0.75) {$-\infty$};
		\node [style=none] (11) at (-3, 1.5) {$0$};
		\node [style=none] (12) at (6, 0.75) {$A_{-}$};
		\node [style=none] (13) at (5.25, 1.5) {$A_+$};
		\node [style=none] (14) at (-1.75, 1.5) {$A$};
		\node [style=none] (15) at (-3.75, 1.5) {$A$};
		\node [style=none] (16) at (1.5, 1.25) {};
		\node [style=none] (17) at (2.5, 1.25) {};
	\end{pgfonlayer}
	\begin{pgfonlayer}{edgelayer}
		\draw [style=newstyle6] (0) to (1.center);
		\draw [style=newstyle6] (0) to (2.center);
		\draw [style=newstyle6] (4) to (3.center);
		\draw [style=newstyle5, bend right=15, looseness=1.50] (4) to (5.center);
	\end{pgfonlayer}
\end{tikzpicture}

%% file: 4-implementation.tex
\section{Implementation}
\label{sec-impl}

In this section, we describe the main components of our
implementation. We first present the transformation framework, where
\TDLLite\ KBs are mapped to \LTL\ formulas, and then we describe the
graphical web tool for generating $\mathcal{ER_{VT}}$ temporal
diagrams (see~\cite{AKRZ:ER10} for details on mapping
$\mathcal{ER_{VT}}$ to a \TDLLite\ TBox, we discuss this more in
Subsection~\ref{sec:wb}).
The suite of tools involved in this framework is composed by a core
Java library, a visual tool for $\mathcal{ER_{VT}}$ diagrams and an
API to connect them. All of the tools are open source and are
available at
\url{https://bitbucket.org/gilia/workspace/projects/TCROW}.

\subsection{The Transformation Framework}

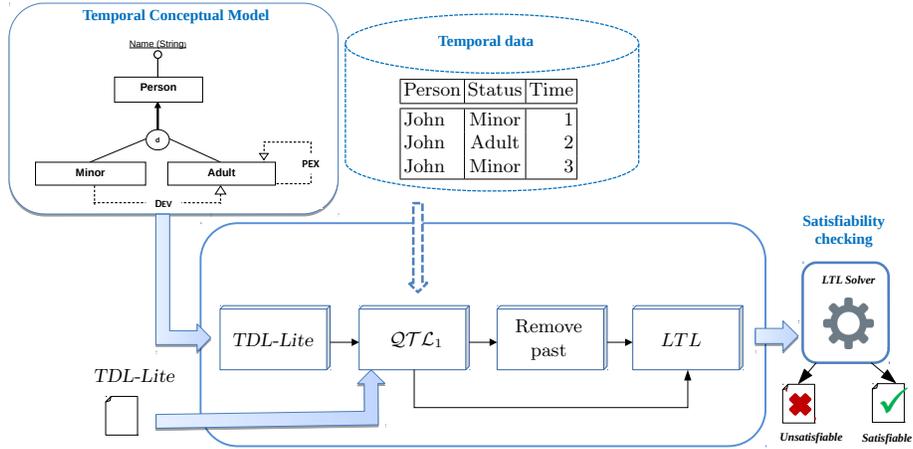
\begin{figure}[t]
 \centering
 \begin{tikzpicture}[scale=0.8, transform shape]
   every node/.style={anchor=south west,inner sep=0pt},
        x=1mm, y=1mm,
      ]   
     \node (fig1) at (0,0)
       {\includegraphics[scale=0.6]{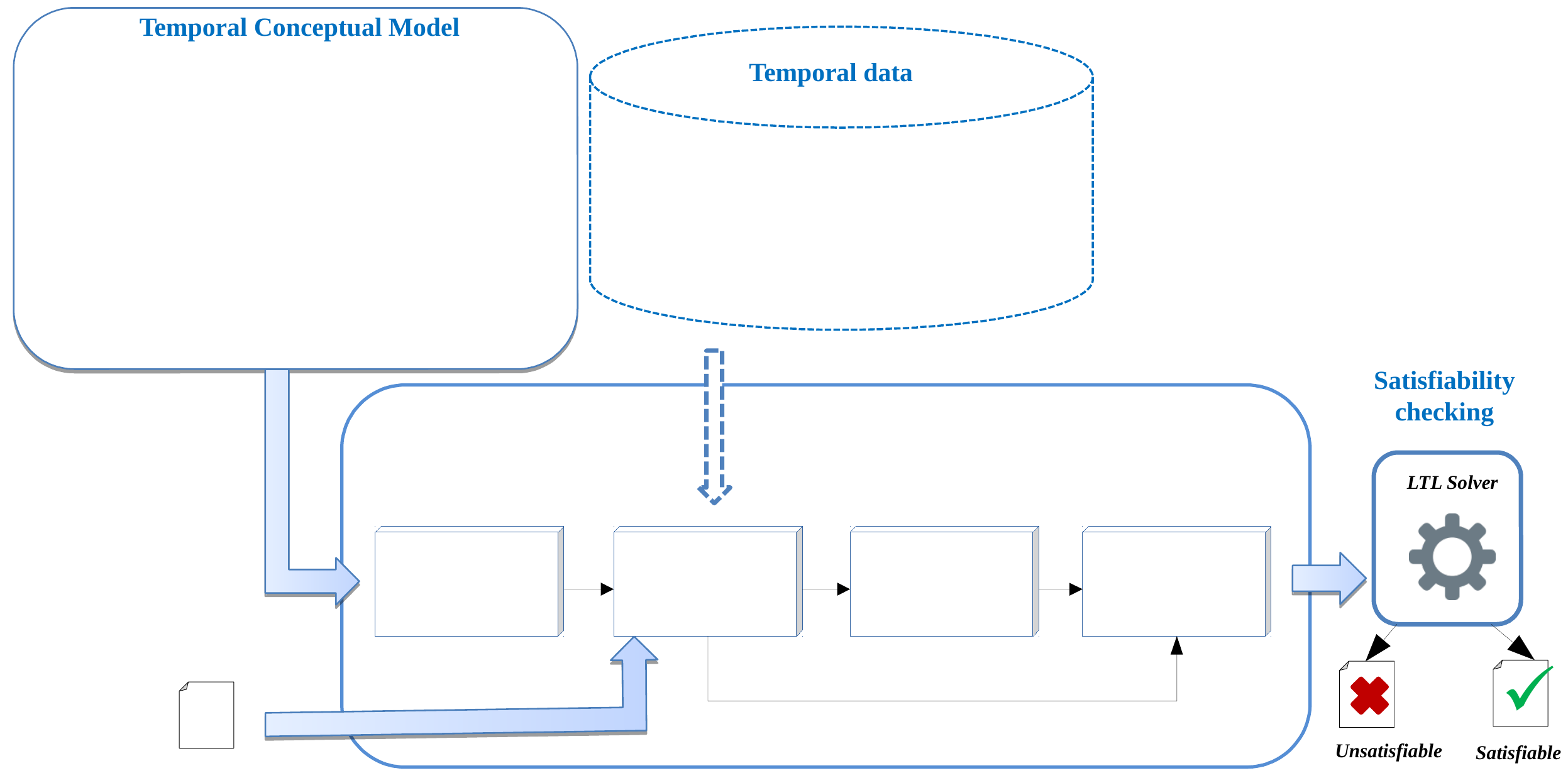}};
     \node (fig2) at (-4.7,1.7)
       {\includegraphics[scale=0.33]{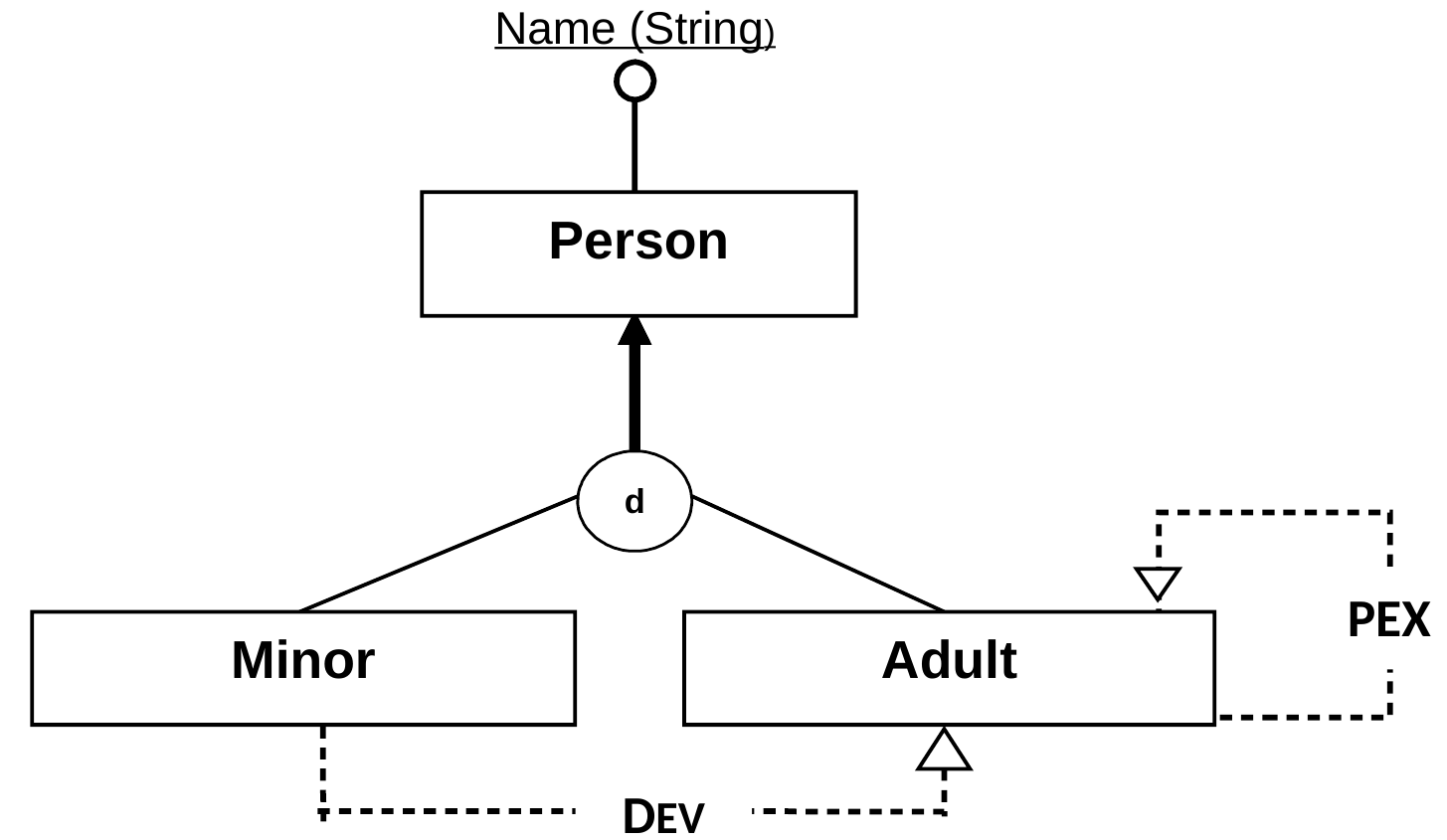}};  
    \draw (-3.1, -1.9) node {\small{\TDLLite}}; 
 \draw (-5.4, -2.5) node {\small{\TDLLite}};
    \draw (-0.7, -1.9) node {\small{\QTLO}};
   \draw (1.5, -1.9) node {
\begin{tabular}{ c }
        Remove \\
        past
\end{tabular}        };
    \draw (3.7, -1.9) node {$LTL$};
\draw (0.5,1.6) node {\begin{tabular}{ |l| c| r| }
\hline
Person & Status & Time\\
\hline
\hline
  John & Minor & 1 \\
  John & Adult & 2 \\
    John & Minor & 3 \\
\hline
\end{tabular}};
\end{tikzpicture}
  \caption{Translation Steps}
 \label{fig:framework}
\end{figure}

Here we comment on our framework.  As shown in
Figure~\ref{fig:framework}, by using our suite of tools one can reason
over \TDLLite\ KBs. We consider two forms of input.  In the first one,
the user draws a temporal conceptual schema using the
$\mathcal{ER_{VT}}$ modelling language which is then automatically
mapped into a \TDLLite TBox.  The graphical interface also allows users to insert temporal data, which is mapped into an ABox.  In the
second, the tool receives as input a \TDLLite KB directly within the
Java code.  Since\nb{A: changed} the OWL language does not support the temporal
dimension, we have created 
ad-hoc methods for 
capturing \TDLLite
KBs 
in the tool.  As mentioned in the Introduction, the tool contains
an implementation of the translation of \TDLLite\ KBs into \LTL\
formulas (Section~\ref{sec-translation}).  This allows the use of
off-the-shelf \LTL\ solvers to perform satisfiability checks for
\TDLLite\ KBs.  The translation is a process with (at most) five main
stages:
\begin{enumerate}
\item Translate the input $\mathcal{ER_{VT}}$ temporal diagram into
  a \TDLLite\ TBox;
\item Mapping the resulting TBox (with the possible addition of ABox
  assertions) into a \QTLO formula (as in Section~\ref{QTL-reduction});
\item Remove the past operators (see Section~\ref{Future-reduction});
\item Obtain the \LTL\ translation via the grounding
  (see Section~\ref{PTL-reduction}); and
\item Adapt\nb{A: delete ``rewrite'' as suggested by REV2} the
  resulting \LTL formula according to the syntax of the corresponding
  solver.
\end{enumerate}

There are two shortcuts for this process
(see~Figure~\ref{fig:framework}).  The first is when the \TDLLite\ KB
is encoded directly within the Java code.  The second is when the
ontology does not contain past operators, i.e., it is expressed is the
\TDLLiteN logic.  While\nb{A: small changes} for clarity of
presentation, Section~\ref{Future-reduction} describes the removal of
past modalities as a translation from \PTL to \LTL, in the current
implementation we remove past operators already from the \QTLO
encoding and then grounding the resulting formula, still obtaining an
equisatisfiable \LTL translation.

\subsection{The $\mathcal{ER_{VT}}$ Web Tool}
\label{sec:wb} 

Involving knowledge engineers, domain experts and end-users in the
construction, maintenance, and use of conceptual models requires a
strong interaction between them. The use of common languages along
with appropriate methodologies is a way to effectively manage such
interactions and to successfully guide the whole process.  However,
methodologies for conceptual modelling appear to be fragmented across
several tools and workarounds \cite{VigoBJS14}, and visualisation
methods proposed are weakly integrated to logic-based reasoning
tools~\cite{DBLP:conf/jowo/BraunCF19}.

Based on these motivations, we have developed an open source graphical
tool for designing $\mathcal{ER_{VT}}$ temporal
diagrams~\cite{DBLP:conf/er/ArtaleF99,artale:franconi:john09,AKRZ:ER10}, named
{\tt
  crowd-$\mathcal{ER_{VT}}$}\footnote{\url{http://crowd.fi.uncoma.edu.ar/ervt-gui/erd_editor.php}}
and based on the architecture presented in~\cite{Braun-KI}.  The idea
behinds the tool is: (i) to define a suitable abstraction level for
temporal modelling tasks involving both users and the underlying
methodologies of conceptual data modelling; and (ii) to understand better how 
logic-based knowledge representation systems can be integrated with
visual languages in a tool.

\begin{figure}[t]
\centering
	\begin{minipage}[b]{0.55\linewidth}
		\centering
		\includegraphics[width=\textwidth]{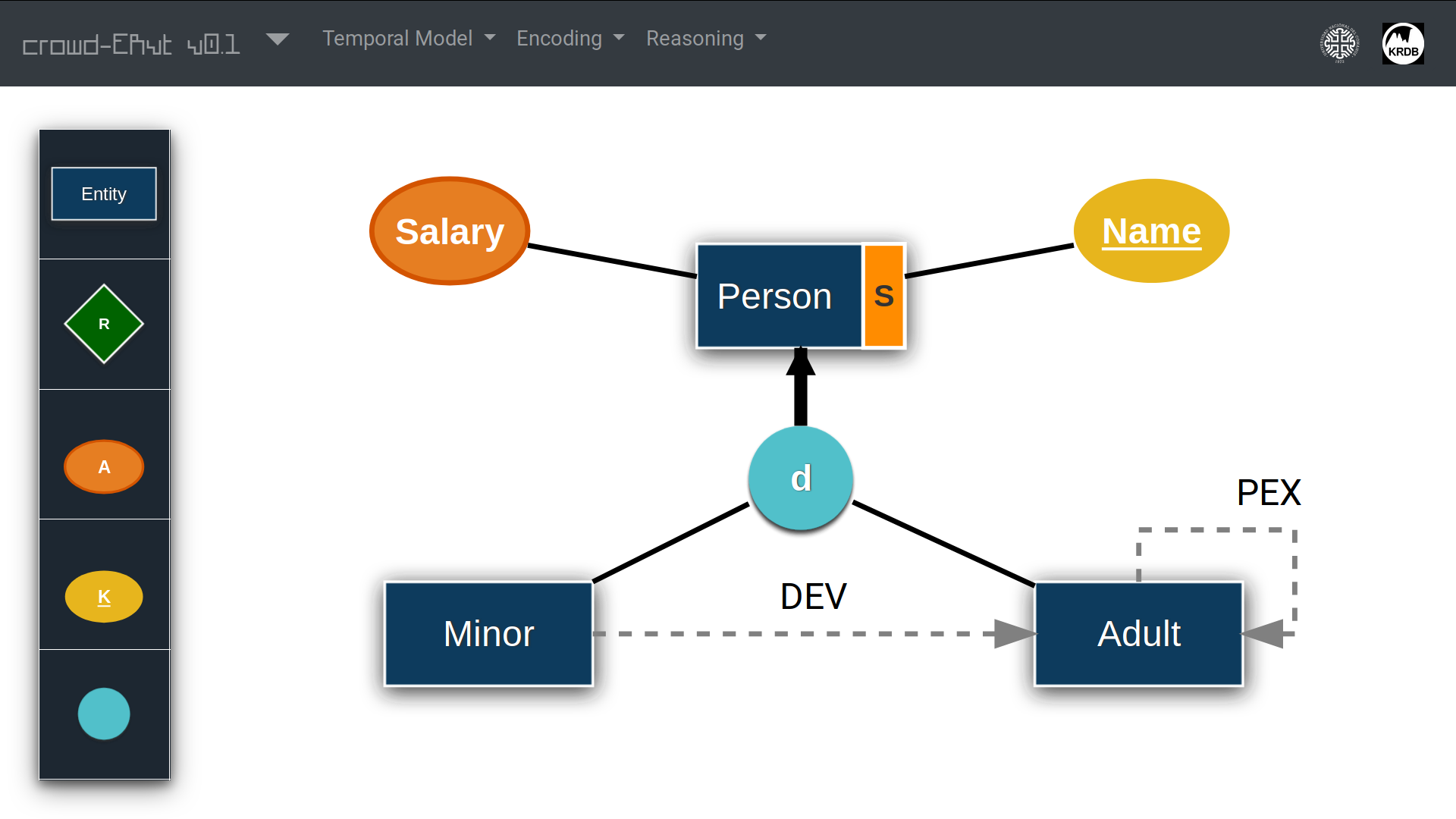}
		\caption{$\mathcal{ER_{VT}}$ graphical interface}
		\label{fig:figure1}
	\end{minipage}
\hspace{0.2cm}
	\begin{minipage}[b]{0.35\linewidth}
		\centering
		\includegraphics[width=\textwidth]{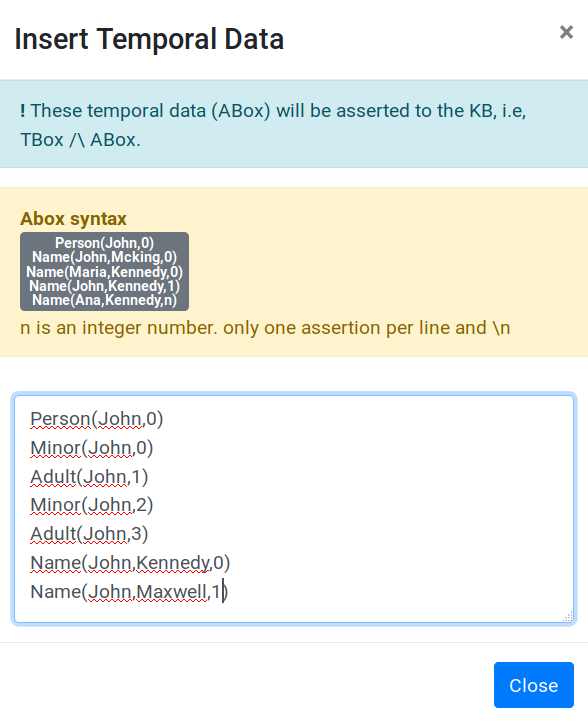}
		\caption{Widget to add temporal  data}
		\label{fig:figure2}
	\end{minipage}
\end{figure}

{\tt crowd-$\mathcal{ER_{VT}}$} supports temporal conceptual modelling
in $\mathcal{ER_{VT}}$ providing a visual environment, a comprehensive
set of $\mathcal{ER_{VT}}$ primitives, and an interface to
off-the-shelf reasoners as depicted in Figure~\ref{fig:framework}.
Moreover, the tool provides an interface to add temporal data
assertions (i.e., a temporal ABox) to populate the $\mathcal{ER_{VT}}$
model. Once both the $\mathcal{ER_{VT}}$ diagram and the temporal data
have been specified, they are jointly sent to the server, where they
are ultimately translated into \LTL.  When the reasoning process ends,
the tool provides a link where the user can download the intermediate
translations, including the \TDLLite\ TBox and ABox encodings, the
respective \QTLO translations, the encoding where past operators have
been removed (if any), the final $\LTL$ formula, and the output of a reasoner.

To show how the tool works, we revisit the running examples introduced
in Section~\ref{ex:1} showing the corresponding $\mathcal{ER_{VT}}$
diagram, and the temporal data assertions.  Figures~\ref{fig:figure1}
and~\ref{fig:figure2} show the graphical interface extending
Examples~\ref{ex:1}-\ref{ex:2}. In addition to the entities {\sf
  Person}, {\sf Adult} and {\sf Minor}, and the
attribute\footnote{\emph{Attributes} in CMs associate an entity to a
  concrete value and are usually encoded as roles in the corresponding
  DL TBox encoding.}\nb{A: added footnote} {\sf Name}, the tool allows to specify \emph{visual marks}
for each temporal primitive in $\mathcal{ER_{VT}}$. For example, {\sf
  Person} is set as a \emph{global} entity (using the {\sc S} mark,
standing for \emph{Snapshot} following the temporal ER
notation~\cite{AKRZ:ER10} for global entities), while each {\sf Minor}
will become an {\sf Adult} sometime in the future (expressed by the
dotted line labeled \texttt{DEV}, standing for Dynamic EVolution). The
axiom stating that minors and adults are disjoint is modelled as a
disjoint composed ISA (the {\bf d} labeled circle). The fact that
every adult stays always adult in the future is captured by the dotted
line labeled \texttt{PEX} (standing for Persistence).
In Figure~\ref{fig:figure2}, we show the user interface to specify temporal
assertions, showing the person {\sf John} and his name at different
timestamps.

The set of visual primitives also includes binary relationships (both
\emph{global} and \emph{local} relationships are supported), and
cardinalities are limited to the
$1..1$ and $0..N$ cases.
Cardinalities on attributes are limited to just $1..1$. Entities and
attributes can be also tagged as \emph{global} or \emph{local}. Lastly,
other forms of temporal operators between entities are also supported: \texttt{TEX} (standing for Transition EXtension) and
\texttt{DEX$^{-}$} (standing for Dynamic EXtension) (see~\cite{DBLP:conf/er/ArtaleF99,artale:franconi:john09,AKRZ:ER10}).



  

%% file: 5-evaluation.tex
\section{Evaluation}
\label{sec-eval}


Since
there are no \TDLLite ontologies available for testing, in our
evaluation, we generated test ontologies synthetically, by extending
methods applied
in the context of propositional temporal logic~\cite{Daniele1999ImprovedAG}.
We describe the test method in Subsection~\ref{subsec:test} and
our results in Subsection~\ref{subsec:results}.
 


\subsection{The Test Method}\label{subsec:test}

We extend the test method proposed by Daniele, Giunchiglia, and
Vardi~\cite{Daniele1999ImprovedAG} in the context of propositional
temporal logic to our case.  We analyse basic properties of the
translation of \TDLLite TBoxes into propositional temporal logic
formulas.  Following the mentioned method, we present two kinds of
analyses.
\begin{itemize}
\item \textbf{Average-behaviour}: For a fixed number of $N$ concept
  names and of $N$ role names, a fixed maximum \MaxQ for the value $q$
  in basic concepts of the form $ \geq q R$, a fixed number
  \lengthTbox of CIs in a TBox, and for increasing values
  \lengthConcept of the length of concept expressions in the TBox, we
  create batches $B(F,N,\lengthTbox,\lengthConcept,\MaxQ)$ of $F$
  random TBoxes. We can then plot the results against \lengthConcept
  and repeat the process for different values of $N$, \MaxQ, and
  \lengthTbox.
\item \textbf{Temporal-behaviour}: For a fixed set of parameters
  $N, \MaxQ, \lengthTbox, \lengthConcept$, and for increasing values
  of the probability $P_t$ of generating the temporal operators
  $\{\D_P, \D_F, \B_P, \B_F\}$\footnote{For the temporal-behaviour, we focused on $\D_{P/F}$ and $\B_{P/F}$
  (and not in $\Next_{P/F}$) to analyse the behaviour several steps ahead of the current time point (as in~\cite{Daniele1999ImprovedAG}). } 
  and the
  probability $P_g$ of generating global roles, we create batches
  $B(F,N,\lengthTbox,\lengthConcept,\MaxQ,P_t,P_g)$ of $F$ random
  TBoxes. We can then plot the results against $P_t$ and $P_g$ and
  repeat the process for different values of $N$, \MaxQ,
  \lengthConcept, and \lengthTbox.
\end{itemize}

We now describe the procedure for generating random \TDLLite TBoxes
drawn uniformly from the space of \TDLLite TBoxes with \lengthTbox CIs
and concept expressions with length \lengthConcept (exactly
\lengthTbox and \lengthConcept, not up to \lengthTbox and
\lengthConcept). These concept expressions are formulated with concept
and role names in $\{A_1,\ldots,A_N\}$ and $\{R_1,\ldots,R_N\}$,
respectively, and $q\leq\MaxQ$ in basic concepts of the form
$ \geq q R$.  Since there is no risk of confusion, in what follows we
omit ``uniformly''.

For the average-behaviour analysis, we proceed as follows.  A random
concept expression with length one is generated by randomly choosing a
basic concept. Random concept expressions of length two are of the
form ${\sf op}(C)$ where $C$ is a random concept expression with
length one and ${\sf op}$ is randomly chosen from
$U=\{\neg, \Next_F,\Next_P, \Diamond_F, \Diamond_P,\Box_F, \Box_P\}$.
For concept expressions with length $L$ larger than one, ${\sf op}$ is
randomly chosen from $U\cup\{\sqcap\}$.  If ${\sf op}\in U$ (that is,
${\sf op}$ is unary) then the random concept expression is
${\sf op} (C)$, where $C$ is a random concept expression with length
$L-1$.  Otherwise, we randomly choose $K$ in $\{1, \ldots, L-2\}$ and
the random concept expression is $C\sqcap D$, where $C$, $D$ are
random concept expressions with length $K$ and $L-K-1$, respectively.
For the temporal-behaviour analysis we use the parameter $P_t$ to
increase the chance of generating the temporal operators in
$\{ \Diamond_F, \Diamond_P,\Box_F, \Box_P\}$ and the parameter $P_r$
increase the chance of generating a global role.  Random concept
expressions with length one are generated as for the average-behaviour
analysis.  For concept expressions with length two, the concept
expression is of the form ${\sf op}(C)$, where the probability that
${\sf op}=\Diamond_F$ is $\frac{P_t}{4}$ (same for ${\sf op}$ being
$\Diamond_P,\Box_F$ or $ \Box_P$) and the probability that
${\sf op}=\neg$ is $\frac{1-P_t}{3}$ (same for the other two unary
operators).  For concept expressions with length greater than two,
${\sf op}$ is again chosen from $U\cup\{\sqcap\}$ but now we
have 
$\frac{1-P_t}{4}$ for the probability that ${\sf op}$ is $\neg$ (same
for $\Next_F,\Next_P,\sqcap$).
We have also generated ABoxes randomly, using the concept and role names
occurring in TBoxes.

We are interested in investigating how different logical constructs
of \TDLLite TBoxes affect the size of the translation, the time for
computing the translation, and the time for deciding satisfiability 
using tools designed for LTL. 
In the temporal-behaviour analysis, we want to study
how the presence of temporal operators
$\Diamond_F, \Diamond_P,\Box_F, \Box_P$ and global roles affect the
size of the translation and the time to compute the translation and to
check for satisfiability.

\subsection{Experimental environment and \LTL solvers}

We briefly present the hardware/software environment where the tests
have been performed together with the off-the-shelf \LTL solvers used.

\paragraph{\bf Hardware and Software.} We run the experiments 
on a server with 8 Genuine Intel 3.6 GHz processors, and 4 GB memory
running Debian GNU/Linux 10 (buster) with 64 bit kernel 4.19.0-8.  Run
time and memory usage were measured with
runlim\footnote{\url{http://fmv.jku.at/runlim/}}.



We set the run time to a timeout of 600 CPU seconds, while memory is set up to 1 GB.
In the test, we denote with `T/O' the case where the solver either
runs out of memory or it times out. With `Fail' we denote the case
where the solver encountered an unexpected error condition.
%

\paragraph{\bf \LTL Solvers.} We considered various \LTL solvers
adopting different reasoning strategies:
\emph{reduction to model checking}, \emph{tableau-based algorithms},
and \emph{temporal resolution}. The chosen solvers have been already
considered in the literature to test their performances for checking \LTL
formula satisfiability~\cite{SchuppanD11,DBLP:conf/cade/HustadtOD17}.
%
As for the solvers based on model checking, we opted for
\texttt{NuXMV} 
using the following options: BDDs (Binary Decision Diagrams), SAT
(based\nb{A: changed} on SAT techniques and also denoted as BMC for
Bounded Model Checking), and the variant IC3 (known also as
\emph{Property Directed
  Reachability})~\cite{nusmv:Cimatti02,nuxmv:CavadaCDGMMMRT14,ic3}.
BDD is a complete satisfiability technique but it requires space
exponential in the number of variables, and it is also sensitive to
the variable order. On the other hand, BMC reduces the problem to SAT
by testing the formula for satisfiability in models with length
bounded by a fixed\nb{A: added} $k$, 
giving sound but incomplete results. 
IC3 computes approximations of reachability in a backward-search
fashion.

The solver \texttt{pltl}~\cite{pltl:98} implements a tableau-based
algorithms. We run \texttt{pltl} with the options \texttt{graph} and
\texttt{tree}. 
Both methods build rooted trees with ancestor cycles but different search strategies (see also~\cite{pltl:graph} for details).  
Compared with the traditional
tableau-based translation, Aalta (Another Algorithm for LTL To Büchi
Automata) \cite{Aalta:19}, implements a new algorithm by introducing
the concept of \emph{Obligation Set} for LTL formulas. Aalta
integrates two functions: LTL-to-Büchi translation and LTL
satisfiability checking.
Finally, \texttt{TRP++UC}~\cite{trpuc:Schuppan13} implements temporal
resolution, and has been used with the option dfs/bfs (Depth-First
search/Breadth-First search).
This solver has been built on 
\texttt{TRP++} \cite{TRP:HustadtK03}. Despite the aim of \texttt{TRP++UC} is to extract
unsatisfiable cores for \LTL, we used the legacy features
of \texttt{TRP++} by running the solver with the unsatisfiable cores
option disabled.

\subsection{Results}
\label{subsec:results}

In what follows we evaluate the efficiency and the scalability of our
tool using two evaluation tests of increasing difficulties.
The first test is based on the \textbf{toy scenarios} described in
Examples~\ref{ex:1} and~\ref{ex:2}. The second test
applies the test method for \textbf{generating TBoxes randomly}, as described in Subsection~\ref{subsec:test}. 
Our benchmark is available from \url{http://crowd.fi.uncoma.edu.ar/temporalDLLite/benchmark/}.
%
We discovered in our experimental evaluation that \texttt{Aalta} presents limitations when the \LTL input has more than 1200 propositional variables. 


\input{5.1-toyexampletest}


%% file: 5.1-toyexampletest.tex
\newcommand{\PreserveBackslash}[1]{\let\temp=\\#1\let\\=\temp}
\newcolumntype{R}[1]{>{\PreserveBackslash\raggedleft}p{#1}}
%
\subsubsection{Toy Scenarios Experiment.}
\label{test:toy examples}
We report here on the experimental results conducted by pairing the
TBoxes of the toy examples, as reported in Section~\ref{sec:intro},
with different ABoxes which may yield satisfiable (SAT) and
unsatisfiable (UNSAT) KBs.  The sizes of the ABoxes vary from $20$ to
$50$ assertions (distributed over different time points).  The number
of propositional variables in the resulting \LTL formula starts from a
minimum of 14 variables, for the KB as in Example~\ref{ex:1}\nb{A:
  added}, and ranges from 180 to 2336 variables for
Example~\ref{ex:2}\nb{A: added} depending on the increasing sizes of
the tested ABoxes.  The experimental results for the cases where the
TBox is as in Examples~\ref{ex:1} and~\ref{ex:2} are shown in
Figures~\ref{Adult-Heat} and \ref{Name-Heat}, respectively.  For\nb{A:
  chenged} the
UNSAT cases, we explicitly inserted 
inconsistent assertions 
(e.g., {\sf John} being both {\sf Minor} and {\sf
  Adult}, 
in case of Example~\ref{ex:1}, while we added more than 1 name to the
same person in case of Example~\ref{ex:2}).
The results, presented in the form of `heat maps', represent\nb{A:
  changed} the runtime of the KB satisfiability checking for
increasing ABox sizes (in columns) and different solvers (in lines).
%
\begin{figure}[t]
  \renewcommand{\arraystretch}{0}
  \setlength{\fboxsep}{0mm}
  \setlength{\tabcolsep}{0pt}
  \begin{center}
    \settowidth{\gapw}{bTD}%
    \begin{tabular}[t]{@{}ll@{\quad}l@{}}
      & \multicolumn{1}{c}{SAT } 
      & \multicolumn{1}{c}{UNSAT }\\[1ex]
      \begin{tabular}{lL}
        \input{Adult-Labels.heat.tex}
      \end{tabular}
      &
        \begin{tabular}{l*{4}{L!{\color{white}\vline}}}
          \input{Adult-Sat.heat.tex}
        \end{tabular}
      &
        \begin{tabular}{l*{4}{L!{\color{white}\vline}}}
          \input{Adult-Unsat.heat.tex}
        \end{tabular}\\
    \end{tabular}\vspace*{-1.5em}    
  \end{center}
  \caption[Heat map for example1]{%
    Heat map with the runtimes for Example~\ref{ex:1}.
    Each rectangle represents the runtime in CPU seconds of
    SAT and UNSAT KBs 
    with ABoxes of increasing size.
    The runtimes are given in colours as follows:\\
    {\scalebox{0.77}{%
\setlength{\fboxsep}{0mm}
\begin{tabular}{@{}*{13}{N@{\ }l@{\ }}@{}}
       0 & $<\lmt{0.01}$ sec &
       1 & $>\lmt{0.01}$ sec, $\leq\lmt{0.25}$ sec &
       2 & $>\lmt{0.25}$ sec, $\leq\lmt{0.50}$ sec &
       3 & $>\lmt{0.50}$ sec, $\leq\lmt{1}$ sec
             \\
       4 & $>\lmt{1}$ sec, $\leq\lmt{2}$ sec &
       5 & $>\lmt{2}$ sec, $\leq\lmt{4}$ sec &
       6 & $>\lmt{4}$ sec, $\leq\lmt{8}$ sec &
       7 & $>\lmt{8}$ sec, $\leq\lmt{16}$ sec 
                                        \\
       8 & $>\lmt{16}$ sec, $\leq\lmt{32}$ sec &
       9 & $>\lmt{32}$ sec, $\leq\lmt{64}$ sec &
      10 & $>\lmt{64}$ sec, $\leq\lmt{125}$ sec &
      11 & $>\lmt{125}$ sec, $\leq\lmt{250}$ sec 
            \\
      12 & $>\lmt{250}$ sec, $\leq\lmt{500}$ sec &
      13 & $>\lmt{500}$ sec, $\leq\lmt{1000}$ sec &
      14 & T/O, OoM or Fail
      \end{tabular}
    }}}\label{Adult-Heat}
\end{figure}%
%
\begin{figure}[t]
  \renewcommand{\arraystretch}{0}
  \setlength{\fboxsep}{0mm}
  \setlength{\tabcolsep}{0pt}
  \begin{center}
    \settowidth{\gapw}{bTD}%
    \begin{tabular}[t]{@{}ll@{\quad}l@{}}
      & \multicolumn{1}{c}{SAT }
      & \multicolumn{1}{c}{UNSAT } \\[1ex]
      \begin{tabular}{lL}
        \input{Name-Labels.heat.tex}
      \end{tabular}
      &
        \begin{tabular}{l*{4}{L!{\color{white}\vline}}}
          \input{Name-Sat.heat.tex}
        \end{tabular}
      &
        \begin{tabular}{l*{4}{L!{\color{white}\vline}}}
          \input{Name-Unsat.heat.tex}
        \end{tabular}\\   
    \end{tabular}\vspace*{-1.5em}
  \end{center}
  \caption[Heat map for Name]{%
  Heat map with the runtimes of Example~\ref{ex:2}.
  Each rectangle represents the runtime in CPU seconds of
  SAT and UNSAT KBs  with ABoxes of increasing size
    (the colours are as in the previous figure).\\
    {\scalebox{0.77}{%
        \setlength{\fboxsep}{0mm}
      }}}\label{Name-Heat}
\end{figure}

Solvers\nb{A: various changes in this paragraph} had better
performances over SAT instances compared to UNSAT ones, except \TRPpp
which fails to scale even to small ABoxes.  Moreover,
\texttt{NuXMV-SBMC} fails regardless the size of the model in UNSAT
cases (also when running in the simpler scenario of
Example~\ref{ex:1}).
Considering the experimental setting of Example~\ref{ex:1}
(Fig.~\ref{Adult-Heat}), we note that the best performers over UNSAT
cases were \texttt{NuXMV} with BDD and \texttt{NuXMV} with
IC3, \texttt{Aalta} and \texttt{pltl}.
The more involved setting of Example~\ref{ex:2} clearly shows that the
BMC option is the best for the SAT cases while \texttt{NuXMV} with the
IC3 option stands out in the UNSAT case.
\texttt{Aalta} performs well but only when the \LTL input does not exceed
$1200$ variables.
%

\begin{figure}[t]
  \renewcommand{\arraystretch}{0}
  \setlength{\fboxsep}{0mm}
  \setlength{\tabcolsep}{0pt}
  \begin{center}
    \settowidth{\gapw}{bTD}%
    \begin{tabular}[t]{@{}llllll@{\quad}l@{}}
       & \multicolumn{1}{c}\texttt{1 concept}
      & \multicolumn{1}{c}\texttt{3 concepts}
      & \multicolumn{1}{c}\texttt{5 concepts}\\  
      \begin{tabular}{lL}
        \input{1average-Labels.heat.tex}
      \end{tabular}
      &
        \begin{tabular}{l*{10}{L}}
          \input{1average-Sat.heat1.tex}
        \end{tabular}
      &
        \begin{tabular}{l*{4}{L!{\color{white}\vline}}}
          \input{1average-Sat.heat2.tex}
        \end{tabular}
      &
        \begin{tabular}{l*{4}{L!{\color{white}\vline}}}
          \input{1average-Sat.heat3.tex}
        \end{tabular}
    &
        \begin{tabular}{l*{4}{L!{\color{white}\vline}}}
          \input{Labels.heatRight.tex}
        \end{tabular}
           \end{tabular}\vspace*{-1.5em}
  \end{center}
  \caption[Heat map for average behaviour]{%
    Heat map of the runtimes on randomly generated TBoxes according to
    the average behaviour (the colors are as in the toy experiments).}
  \label{average-Heat}
\end{figure}
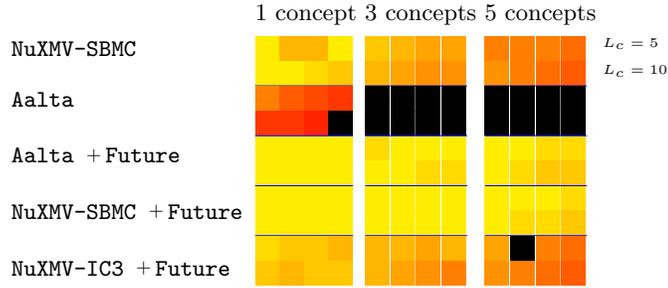
\begin{figure}[t]
  \renewcommand{\arraystretch}{0}
  \setlength{\fboxsep}{0mm}
  \setlength{\tabcolsep}{0pt}
  \begin{center}
    \settowidth{\gapw}{bTD}%
    \begin{tabular}[t]{@{}llllll@{\quad}l@{}}
       & \multicolumn{1}{c}\texttt{1 concept}
      & \multicolumn{1}{c}\texttt{3 concepts}
      & \multicolumn{1}{c}\texttt{5 concepts}\\  
      \begin{tabular}{lL}
        \input{Temporal-Labels.heat.tex}
      \end{tabular}
      &
        \begin{tabular}{l*{4}{L!{\color{white}\vline}}}
          \input{Temporal-Sat.heat1.tex}
        \end{tabular}
      &
        \begin{tabular}{l*{4}{L!{\color{white}\vline}}}
          \input{Temporal-Sat.heat2.tex}
        \end{tabular}
      &
        \begin{tabular}{l*{4}{L!{\color{white}\vline}}}
          \input{Temporal-Sat.heat3.tex}
        \end{tabular}
        &
        \begin{tabular}{l*{4}{L!{\color{white}\vline}}}
          \input{Labels.heatRight.tex}
        \end{tabular}
    \end{tabular}\vspace*{-1.5em}
  \end{center}
  \caption[Heat map for the temporal behaviour]{%
    Heat map of the runtimes on randomly generated TBoxes according to the temporal behaviour (the colors are as in the toy experiments).\\
    {\scalebox{0.77}{%
        \setlength{\fboxsep}{0mm}
      }}}\label{Temporal-Heat}
\end{figure}
\subsubsection{Randomly Generated TBoxes.}
\label{test:synthetic TBoxes}
With this experiment setting we investigate the reasoning scalability
over randomly generated TBoxes (see Section~\ref{subsec:test}). These
are KBs with an empty ABox and with (mostly) SAT TBoxes.  In this is
evaluation we only consider \LTL solvers that performed well in the
previous toy scenarios, namely, \texttt{NuXMV-SBMC}, \texttt{Aalta},
and \texttt{NuXMV-IC3}. Again, we use `heat maps' to illustrate the
obained results. For each solver,\nb{A: added} the results are
reported in two consecutive lines: the first line denoting the case where
$\lengthConcept = 5$, while the second has $\lengthConcept =
10$. Each square represents the runtime for different values of $N$
according to the following legenda:
\begin{itemize}
\item For the average behaviour, TBoxes were generated with the
  parameters $F = 20$, $N = 1,3,5$, $\lengthTbox = 5,10,15,20$
  (columns), $\lengthConcept = 5,10$ (lines), and $\MaxQ = 5$.
\item For the temporal behaviour, TBoxes are generated with the
  parameters $F = 20$, $N = 1,3,5$, $\lengthTbox = 10$,
  $\lengthConcept = 5,10$ (lines), $\MaxQ = 5$, $P_t =0.1,0.9$, and
  $P_g =0.7,0.9 $ (the first two columns are with $P_g=0.7$ and two
  values for $P_t=0.1, 0.9$, the last two columns are with $P_g=0.9$
  and again $P_t=0.1, 0.9$).
\end{itemize}

\nb{G: paragraph added.} Concerning the number of propositional variables in the resulting \LTL formulae and the current experimental setting, 
the average behaviour generated formulae from 400 and up to 8400 variables (\lengthConcept = 5), and
from 500 and up to 14500 variables (\lengthConcept = 10). On the other hand, the
temporal behaviour generated from 400 and up to 3500 variables (\lengthConcept = 5), and
from 950 and up to 6600 variables (\lengthConcept = 10).

Due \nb{A: new sentence} to the increase in the number of variables
when removing the past operators, as expected solvers perform better
on TBoxes expressed with only future operators (i.e., on $\TDLLiteN$
TBoxes) as shown on lines 3, 4 and 5 on both Figures~\ref{average-Heat}
and~\ref{Temporal-Heat}, with the BMC option performing better than
IC3.
%
%
In the general case (presence of both past and future
operators),\nb{A: small changes} the \LTL translation that removes
past operators at the cost of a linear increase of the propositional
variables 
makes the reasoning more difficult.  Moreover, there is not much
difference between the heat map corresponding to the average
behavior (Fig.~\ref{average-Heat}) and the one of the temporal
behavior (Fig.~\ref{Temporal-Heat}). In our experimental
setting, this indicates that the \LTL solvers were more affected by
the number of propositional variables involved in the formula than by
the number of temporal operators in the formula. Once again the BMC
option is the best one for reasoning on SAT cases.
%


%% file: Adult-Labels.heat.tex
\makebox[\gapw][l]{}\Aalta & 20\\

\makebox[\gapw][l]{}\PLTLGraph & 20\\

\makebox[\gapw][l]{}\PLTLTree & 20\\

\makebox[\gapw][l]{}\NuXMVBDD & 20\\

\makebox[\gapw][l]{}\NuXMVSBMC\texttt{60}& 20\\

\makebox[\gapw][l]{}\NuXMVSBMC \texttt{20} & 20\\

\makebox[\gapw][l]{}\NuXMVIC & 20\\

\makebox[\gapw][l]{}\TRPppBFS & 20\\

\makebox[\gapw][l]{}\TRPppDFS & 20\\

%% file: Adult-Sat.heat.tex
&      1&      3&       3&      3\\
\arrayrulecolor{blue}\hline
&      2&      2&      2&      2\\
&      2&      2&      2&      2\\
\arrayrulecolor{blue}\hline
&      2&      2&      2&      2\\
\arrayrulecolor{blue}\hline
&      2&      2&      2&      2\\

&      2&      2&      14&      14\\
\arrayrulecolor{blue}\hline
&      2&      3&      3&      6\\
\arrayrulecolor{blue}\hline
&      5&      14&      14&      14\\
&      6&      14&      14&      14\\
\arrayrulecolor{blue}\hline

%% file: Adult-Unsat.heat.tex
&      1&      2&      2&      2\\
\arrayrulecolor{blue}\hline
&      1&      2&      2&      2\\
&      2&      2&      2&      2\\
\arrayrulecolor{blue}\hline
&      2&      2&      2&      2\\
\arrayrulecolor{blue}\hline
&      14&      14&      14&      14\\

&      14&      14&      14&      14\\
\arrayrulecolor{blue}\hline
&      2&      2&      2&      2\\
\arrayrulecolor{blue}\hline
&      4&      14&      14&      14\\
&      4&      14&      14&      14\\
\arrayrulecolor{blue}\hline

%% file: Name-Labels.heat.tex
\makebox[\gapw][l]{}\Aalta & 20\\
\makebox[\gapw][l]{}\PLTLGraph & 20\\
\makebox[\gapw][l]{}\PLTLTree & 20\\

\makebox[\gapw][l]{}\NuXMVBDD & 20\\

\makebox[\gapw][l]{}\NuXMVSBMC\texttt{60} & 20\\

\makebox[\gapw][l]{}\NuXMVSBMC\texttt{20}& 20\\

\makebox[\gapw][l]{}\NuXMVIC & 20\\

\makebox[\gapw][l]{}\TRPppBFS & 20\\

\makebox[\gapw][l]{}\TRPppDFS & 20\\

%% file: Name-Sat.heat.tex
&      3&      3&      3&      3\\
\arrayrulecolor{blue}\hline

&      14&      14&      14&      14\\

&      14&      14&      14&      14\\
\arrayrulecolor{blue}\hline

&      14&      14&      14&      14\\
\arrayrulecolor{blue}\hline

&      14&      3&      3&      3\\

&      2&      2&      3&      3\\
\arrayrulecolor{blue}\hline

&      4&      3&      3&      3\\
\arrayrulecolor{blue}\hline

&   14&   14&   14&   14\\

&   14&   14&   14&   14\\
\arrayrulecolor{blue}\hline

%% file: Name-Unsat.heat.tex
&      9&      10&      14&      14\\
\arrayrulecolor{blue}\hline

&      14&      14&      14&      14\\

&      14&      14&      14&      14\\
\arrayrulecolor{blue}\hline

&      14&      14&      14&      14\\
\arrayrulecolor{blue}\hline

&      14&      14&      14&      14\\

&      14&      14&      14&      14\\
\arrayrulecolor{blue}\hline

&      3&      4&      4&      5\\
\arrayrulecolor{blue}\hline

&      14&      14&      14&      14\\

&      14&      14&      14&      14\\
\arrayrulecolor{blue}\hline

%% file: 1average-Labels.heat.tex
\makebox[\gapw][l]{}\phantom{ rien}\\
\makebox[\gapw][l]{}\NuXMVSBMC  \\
\makebox[\gapw][l]{}\phantom{ rien}\\
\makebox[\gapw][l]{}\phantom{ rien}\\
\makebox[\gapw][l]{}\Aalta & 20\\
\makebox[\gapw][l]{}\phantom{ rien}\\

\makebox[\gapw][l]{}\phantom{ rien}\\
\makebox[\gapw][l]{}\Aalta\splus\texttt{Future} & 20\\
\makebox[\gapw][l]{}\phantom{ rien}\\

\makebox[\gapw][l]{}\phantom{ rien}\\
\makebox[\gapw][l]{}\NuXMVSBMC\splus\texttt{Future}& 20\\
\makebox[\gapw][l]{}\phantom{ rien}\\

\makebox[\gapw][l]{}\phantom{ rien}\\
\makebox[\gapw][l]{}\NuXMVIC \splus\texttt{Future}  & 20\\
\makebox[\gapw][l]{}\phantom{ rien}\\

%% file: 1average-Sat.heat1.tex
&   1&   4&   4&  1  \\
&   1&   1&   2& 3  \\
\arrayrulecolor{blue}\hline

&   7&   9&   10&  11 \\
&   11&   11&   12& 14   \\
\arrayrulecolor{blue}\hline

&   1&   1&   1&   1\\
&   1&   1&   1&   1\\
\arrayrulecolor{blue}\hline

&   1&   1&   1&  1\\
&   1&   1&   1&  1\\
\arrayrulecolor{blue}\hline
&   2&   3&   3&   4\\
&   3&   4&   3&   3\\

%% file: 1average-Sat.heat2.tex
&   3&   4&   5&  5  \\
&   4&   5&   6&  6  \\
\arrayrulecolor{blue}\hline

&   14&   14&   14&  14 \\
&   14&   14&   14& 14   \\
\arrayrulecolor{blue}\hline

&   2&   1&   1&   1\\
&   1&   1&   2&   2\\
\arrayrulecolor{blue}\hline

&   1&   1&   1&  1\\
&   1&   1&   1&  1\\
\arrayrulecolor{blue}\hline
&   4&   4&   5&   4\\
&   4&   5&   6&   7\\

%% file: 1average-Sat.heat3.tex
&   7&   7&   7&  8  \\
&   6&   7&   8&  9  \\
\arrayrulecolor{blue}\hline

&   14&   14&   14& 14  \\
&   14&   14&   14& 14   \\
\arrayrulecolor{blue}\hline

&   1&   1&   2&   2\\
&   1&   2&   3&   3\\
\arrayrulecolor{blue}\hline

&   1&   1&   1&  2\\
&   1&   2&   2&  3\\
\arrayrulecolor{blue}\hline
&   5&   14&   7&  8 \\
&   6&   7&   8&   9\\

%% file: Labels.heatRight.tex
\tiny{   $\lengthConcept = 5$}\\
\makebox[\gapw][l]{}\phantom{ rien}\\
\tiny{   $\lengthConcept = 10$}\\
\makebox[\gapw][l]{}\phantom{ rien}\\
\makebox[\gapw][l]{}\phantom{ rien}\\
\makebox[\gapw][l]{}\phantom{ rien}\\
\makebox[\gapw][l]{}\phantom{ rien}\\
\makebox[\gapw][l]{}\phantom{ rien}\\
\makebox[\gapw][l]{}\phantom{ rien}\\
\makebox[\gapw][l]{}\phantom{ rien}\\
\makebox[\gapw][l]{}\phantom{ rien}\\
\makebox[\gapw][l]{}\phantom{ rien}\\
\makebox[\gapw][l]{}\phantom{ rien}\\
\makebox[\gapw][l]{}\phantom{ rien}\\
\makebox[\gapw][l]{}\phantom{ rien}\\
\makebox[\gapw][l]{}\phantom{ rien}\\

%% file: Temporal-Labels.heat.tex
\makebox[\gapw][l]{}\phantom{ rien}\\
\makebox[\gapw][l]{}\NuXMVSBMC  \\
\makebox[\gapw][l]{}\phantom{ rien}\\
\makebox[\gapw][l]{}\phantom{ rien}\\
\makebox[\gapw][l]{}\Aalta & 20\\
\makebox[\gapw][l]{}\phantom{ rien}\\

\makebox[\gapw][l]{}\phantom{ rien}\\
\makebox[\gapw][l]{}\Aalta\splus\texttt{Future} & 20\\
\makebox[\gapw][l]{}\phantom{ rien}\\

\makebox[\gapw][l]{}\phantom{ rien}\\
\makebox[\gapw][l]{}\NuXMVSBMC\splus\texttt{Future}& 20\\
\makebox[\gapw][l]{}\phantom{ rien}\\

\makebox[\gapw][l]{}\phantom{ rien}\\
\makebox[\gapw][l]{}\NuXMVIC \splus\texttt{Future}  & 20\\
\makebox[\gapw][l]{}\phantom{ rien}\\

%% file: Temporal-Sat.heat1.tex
&   1&   1&   3&  4  \\
&   1&   1&   2&  2 \\
\arrayrulecolor{blue}\hline
&   8&   10&     11&  12  \\
&   9&   10&   11&  13   \\
\arrayrulecolor{blue}\hline

&   1&   1&   1&  1   \\
&   1&  1&   1&  1 \\
\arrayrulecolor{blue}\hline

&   1&  1&   1&  1\\
&   1&   1&   1&  1 \\
\arrayrulecolor{blue}\hline
&   3&   3&   6&  3 \\
&   3&   3&   4&  4 \\

%% file: Temporal-Sat.heat2.tex
&   3&   3&   4&  4   \\
&   5&   5&   5&  5   \\
\arrayrulecolor{blue}\hline
&   14&   14&  14&  14  \\
&   14&   14&  14&  14  \\
\arrayrulecolor{blue}\hline

&   1&   1&   2&  2  \\
&   1&   1&   2&  2 \\
\arrayrulecolor{blue}\hline

&   1&   1&  1& 1  \\
&   1&   1&  1&  1\\
\arrayrulecolor{blue}\hline
&   4&   4&   5&  5  \\
&   5&   5&   6&  6\\

%% file: Temporal-Sat.heat3.tex
&   5&  5&   5&   5   \\
&   7&   7&   7&  7    \\
\arrayrulecolor{blue}\hline
&   14&   14&   14& 14   \\
&   14&   14&   14&  14    \\
\arrayrulecolor{blue}\hline

&   1&  1&   1& 1   \\
&   1&  1&   1& 1  \\
\arrayrulecolor{blue}\hline

&   1&   2&   2&  2  \\
&   1&   1&   1& 1  \\
\arrayrulecolor{blue}\hline
&   5&   6&   6&  6   \\
&   7&   7&   7&  7  \\

%% file: 6-conclusion.tex
\section{Conclusions}
\label{sec:conclusion}
We illustrated the feasibility of automated reasoning on \TDLLite KBs
by leveraging existing off-the-shelf \LTL reasoners. We tested both
toy examples, where the addition of the ABox makes the toy KBs
unsatisfiable, and randomly generated KBs. While the random tests gave
a measure of the scalability and robustness of the off-the-shelf
solvers while checking TBoxes satisfiability, the\nb{A: new sentence}
setting of the toy scenario was useful to check the behaviours of the
solvers when dealing with ABoxes.

To\nb{A: added paragraph} summerize the discoveries, two major
culprits in the runtime of solvers are the size of the ABox and the
presence of past operators. While reasoners are not able to check
formulas with past operators interpreted over the integers,
eliminating past operators increases the number of propositional
variables in a formula and this penalizes the runtime of the
solvers. Better performances have been obtained when checking KBs with
just future operators. Concerning ABoxes, the preliminary results show
that a brute force approach makes reasoning in the presence of ABoxes
almost unfeasable and different approaches need to be investigated.

Being the first attempt ever to check how \LTL solvers deal with
\TDLLite KBs we foresee various open research directions to improve
the results obtained here. On one side, one could aim at developing
algorithms to deal directly with the DL/first-order case avoiding the
\LTL grounding that makes the size of the resulting \LTL encoding not
manageable, in particular, in the presence of ABoxes. How to deal with
ABoxes is still open and current techniques embedded in DL reasoners
dealing with real size ABox should be investigated. In this
respect,\nb{A: added sentence} we plan to extend to the temporal
case the existing ABox abstraction approaches, which are successfully
applied over OWL ontologies~\cite{FokoueMSP12,GlimmKT16}.


%% file: paper.bbl
\begin{thebibliography}{10}

\bibitem{ArtEtAl1}
A.~Artale, D.~Calvanese, R.~Kontchakov, and M.~Zakharyaschev.
\newblock The {DL-L}ite family and relations.
\newblock {\em J. Artif. Intell. Res. {(JAIR)}}, 36(1):1--69, 2009.

\bibitem{DBLP:conf/er/ArtaleF99}
A.~Artale and E.~Franconi.
\newblock Temporal {ER} modeling with description logics.
\newblock In {\em Conceptual Modeling - {ER} '99, 18th International Conference
  on Conceptual Modeling, Proceedings}, pages 81--95, 1999.

\bibitem{ArtFra2}
A.~Artale and E.~Franconi.
\newblock Temporal description logics.
\newblock In {\em Handbook of Temporal Reasoning in Artificial Intelligence},
  pages 375--388. Elsevier, 2005.

\bibitem{artale:franconi:john09}
A.~Artale and E.~Franconi.
\newblock Foundations of temporal conceptual data models.
\newblock In A.~Borgida, V.~Chaudhri, P.~Giorgini, and E.~Yu, editors, {\em
  Conceptual Modeling: Foundations and Applications}, volume LNCS 5600 of {\em
  Lecture Notes in Computer Science}, pages 10--35. Springer, 2009.
\newblock ISBN: 978-3-642-02462-7.

\bibitem{AKRZ:ER10}
A.~Artale, R.~Kontchakov, V.~Ryzhikov, and M.~Zakharyaschev.
\newblock Complexity of reasoning over temporal data models.
\newblock In {\em Proc.\ of the $29^{th}$ International Conference on
  Conceptual Modeling (ER-10)}, 2010.

\bibitem{ArtEtAl3}
A.~Artale, R.~Kontchakov, V.~Ryzhikov, and M.~Zakharyaschev.
\newblock A cookbook for temporal conceptual data modelling with description
  logics.
\newblock {\em {ACM} Trans. Comput. Log.}, 15(3):25:1--25:50, 2014.

\bibitem{BaaEtAl2}
F.~Baader, S.~Ghilardi, and C.~Lutz.
\newblock {LTL} over description logic axioms.
\newblock {\em {ACM} Trans. Comput. Log.}, 13(3), 2012.

\bibitem{DBLP:conf/jowo/BraunCF19}
G.~A. Braun, L.~A. Cecchi, and P.~R. Fillottrani.
\newblock Taking advantages of automated reasoning in visual ontology
  engineering environments.
\newblock In {\em Proc. \ of the Joint Ontology Workshops 2019 Episode {V:} The
  Styrian Autumn of Ontology (JOWO'19)}, 2019.

\bibitem{Braun-KI}
G.~A. Braun, C.~Gimenez, L.~A. Cecchi, and P.~R. Fillottrani.
\newblock {\tt crowd}: A visual tool for involving stakeholders into ontology
  engineering tasks.
\newblock {\em Künstl Intell (2020)}, 2020.

\bibitem{nuxmv:CavadaCDGMMMRT14}
R.~Cavada, A.~Cimatti, M.~Dorigatti, A.~Griggio, A.~Mariotti, A.~Micheli,
  S.~Mover, M.~Roveri, and S.~Tonetta.
\newblock The {NuXMV} symbolic model checker.
\newblock In {\em 26th Int. Conf. on Computer Aided Verification, {(CAV)}},
  volume 8559 of {\em Lecture Notes in Computer Science}, pages 334--342.
  Springer, 2014.

\bibitem{nusmv:Cimatti02}
A.~Cimatti, E.~Clarke, E.~Giunchiglia, F.~Giunchiglia, M.~Pistore, M.~Roveri,
  R.~Sebastiani, and A.~Tacchella.
\newblock {NuSMV} 2: An opensource tool for symbolic model checking.
\newblock In E.~Brinksma and K.~G. Larsen, editors, {\em Computer Aided
  Verification}, pages 359--364. Springer Berlin Heidelberg, 2002.

\bibitem{Daniele1999ImprovedAG}
M.~Daniele, F.~Giunchiglia, and M.~Y. Vardi.
\newblock Improved automata generation for linear temporal logic.
\newblock In {\em CAV}, 1999.

\bibitem{ic3}
N.~Een, A.~Mishchenko, and R.~Brayton.
\newblock Efficient implementation of property directed reachability.
\newblock In {\em Proc. of the Int. Conf. on Formal Methods in Computer-Aided
  Design}, FMCAD-11, page 125–134, 2011.

\bibitem{FokoueMSP12}
A.~Fokoue, F.~Meneguzzi, M.~Sensoy, and J.~Z. Pan.
\newblock Querying linked ontological data through distributed summarization.
\newblock In J.~Hoffmann and B.~Selman, editors, {\em Proceedings of the
  Twenty-Sixth {AAAI} Conference on Artificial Intelligence}. {AAAI} Press,
  2012.

\bibitem{Gab89}
D.~M. Gabbay.
\newblock The declarative past and imperative future: Executable temporal logic
  for interactive systems.
\newblock In B.~Banieqbal, H.~Barringer, and A.~Pnueli, editors, {\em
  Proceedings of the 1st Conference on Temporal Logic in Specification}, volume
  398 of {\em Lecture Notes in Computer Science}, page 409–448.
  Springer-Verlag, 1989.

\bibitem{Gab80}
D.~M. Gabbay, A.~Pnueli, S.~Shelah, and J.~Stavi.
\newblock On the temporal analysis of fairness.
\newblock In {\em Conference Record of the 7th {ACM} Symposium on Principles of
  Programming Languages (POPL’80)}, page 163–173. ACM Press, 1980.

\bibitem{DBLP:conf/lpar/GiganteMR17}
N.~Gigante, A.~Montanari, and M.~Reynolds.
\newblock A one-pass tree-shaped tableau for {LTL}+past.
\newblock In T.~Eiter and D.~Sands, editors, {\em {LPAR-21}}, volume~46 of {\em
  EPiC Series in Computing}, pages 456--473, 2017.

\bibitem{GlimmKT16}
B.~Glimm, Y.~Kazakov, and T.~Tran.
\newblock Scalable reasoning by abstraction beyond {DL-Lite}.
\newblock In M.~Ortiz and S.~Schlobach, editors, {\em Web Reasoning and Rule
  Systems ({RR}) - Proceedings of the 10th International Conference}, volume
  9898 of {\em Lecture Notes in Computer Science}, pages 77--93. Springer,
  2016.

\bibitem{pltl:graph}
R.~Gor{\'e} and F.~Widmann.
\newblock An optimal on-the-fly tableau-based decision procedure for
  {PDL}-satisfiability.
\newblock In R.~A. Schmidt, editor, {\em Automated Deduction, CADE-22}, pages
  437--452, 2009.

\bibitem{TRP:HustadtK03}
U.~Hustadt and B.~Konev.
\newblock {TRP++2.0:} {A} temporal resolution prover.
\newblock In {\em Automated Deduction - CADE-19, 19th International Conference
  on Automated Deduction, Proceedings}, pages 274--278, 2003.

\bibitem{DBLP:conf/cade/HustadtOD17}
U.~Hustadt, A.~Ozaki, and C.~Dixon.
\newblock Theorem proving for metric temporal logic over the naturals.
\newblock In {\em {CADE}}, pages 326--343, 2017.

\bibitem{Aalta:19}
J.~Li, S.~Zhu, G.~Pu, L.~Zhang, and M.~Y. Vardi.
\newblock Sat-based explicit {LTL} reasoning and its application to
  satisfiability checking.
\newblock {\em Formal Methods Syst. Des.}, 54(2):164--190, 2019.

\bibitem{LutzWZ08}
C.~Lutz, F.~Wolter, and M.~Zakharyaschev.
\newblock Temporal description logics: {A} survey.
\newblock In {\em Proc.\ of the 15th Int. Symposium on Temporal Representation
  and Reasoning, TIME'08}, pages 3--14. {IEEE} Computer Society, 2008.

\bibitem{Markey03}
N.~Markey.
\newblock Temporal logic with past is exponentially more succinct.
\newblock {\em Bulletin of the {EATCS}}, 79:122--128, 2003.

\bibitem{PoggiLCGLR08}
A.~Poggi, D.~Lembo, D.~Calvanese, G.~D. Giacomo, M.~Lenzerini, and R.~Rosati.
\newblock Linking data to ontologies.
\newblock {\em J. of Data Semantics}, 10:133--173, 2008.

\bibitem{trpuc:Schuppan13}
V.~Schuppan.
\newblock Extracting unsatisfiable cores for {LTL} via temporal resolution.
\newblock In C.~S{\'{a}}nchez, K.~B. Venable, and E.~Zim{\'{a}}nyi, editors,
  {\em 2013 20th International Symposium on Temporal Representation and
  Reasoning, Pensacola, FL, USA, September 26-28, 2013}, pages 54--61. {IEEE}
  Computer Society, 2013.

\bibitem{SchuppanD11}
V.~Schuppan and L.~Darmawan.
\newblock Evaluating {LTL} satisfiability solvers.
\newblock In {\em Automated Technology for Verification and Analysis, 9th
  International Symposium, {ATVA} 2011. Proceedings}, pages 397--413, 2011.

\bibitem{pltl:98}
S.~Schwendimann.
\newblock A new one-pass tableau calculus for {PLTL}.
\newblock In H.~de~Swart, editor, {\em Automated Reasoning with Analytic
  Tableaux and Related Methods}, pages 277--291. Springer Berlin Heidelberg,
  1998.

\bibitem{Tseytin66}
G.~Tseytin.
\newblock On the complexity of derivation in propositional calculus.
\newblock In {\em Leningrad Seminar on Mathematical Logic}, 1966.

\bibitem{VigoBJS14}
M.~Vigo, S.~Bail, C.~Jay, and R.~D. Stevens.
\newblock Overcoming the pitfalls of ontology authoring: Strategies and
  implications for tool design.
\newblock {\em Int. J. Hum. Comput. Stud.}, 72(12):835--845, 2014.

\end{thebibliography}
